\documentclass[runningheads]{llncs}

\usepackage{eccv}

\usepackage{eccvabbrv}

\usepackage{graphicx}
\usepackage{booktabs}
\usepackage{multicol}
\usepackage{pifont}
\usepackage{multirow}
\usepackage{xcolor}
\usepackage{colortbl}
\usepackage{makecell}
\usepackage{wrapfig}
\usepackage{caption}
\usepackage[accsupp]{axessibility}  
\newcommand{\cmark}{\ding{51}}
\newcommand{\xmark}{\ding{55}}

\newcommand{\modelname}[1]{ConsiSpace}

\usepackage{hyperref}

\usepackage{orcidlink}

\begin{document}

\title{\modelname{}: Learning Geometric Consistency Matters for Video Spatial Reasoning} 

\titlerunning{\modelname{}}

\author{Ting Huang\inst{2}\orcidlink{0009-0004-5042-0908} \and
Zhenyu Zhang\inst{1}$^\dag$ \and
Wenyuan Huang\inst{1} \and \\
Jian Yang\inst{1} \and
Hao Tang\inst{2,3}$^\dag$\orcidlink{0000-0002-2077-1246}
} 

\authorrunning{T.~Huang et al.}

\institute{
School of Intelligent Science and Technology, Nanjing University, Nanjing, China  \and
School of Computer Science, Peking University, Beijing, China  \and
Beijing Academy of Artificial Intelligence, Beijing, China \\
$^\dag$Corresponding authors: \email{zhangjesse@foxmail.com bjdxtanghao@gmail.com}.\\
\url{https://believeht029.github.io/ConsiSpace/}
}
 
\makeatletter
\let\@oldmaketitle\@maketitle%
\renewcommand{\@maketitle}{\@oldmaketitle%
{
\centering
\includegraphics[width=\linewidth]{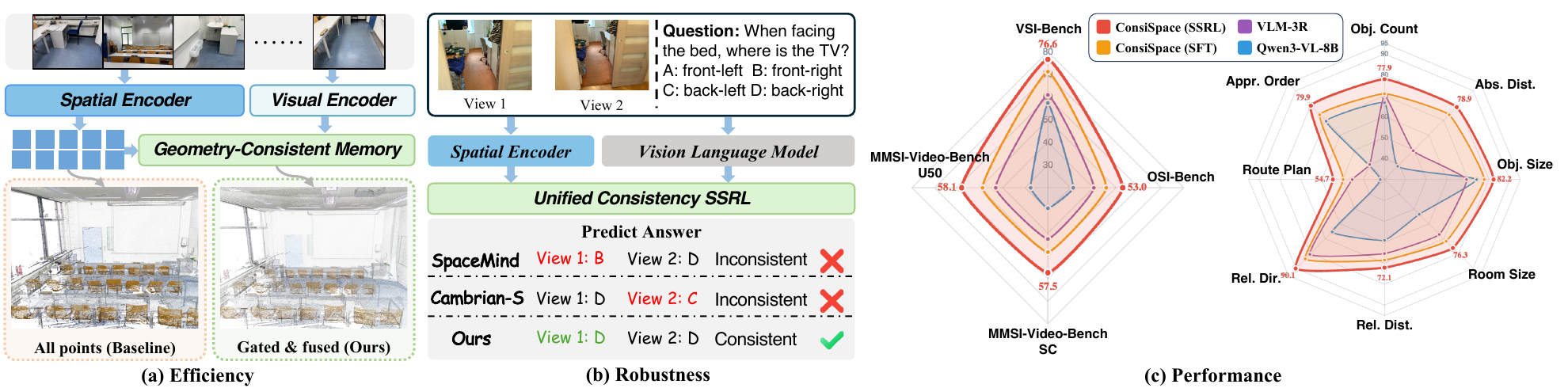}
\small
\captionof{figure}{\textbf{\modelname{} learning geometric consistency for video spatial reasoning.}
\textbf{(a) Efficiency:} Geometry-consistent memory reduces redundant spatial evidence through gated writing and fusion.
\textbf{(b) Robustness:} Unified Consistency SSRL improves answer stability across viewpoints.
\textbf{(c) Performance:} \modelname{} consistently improves spatial reasoning performance across benchmarks and VSI sub-tasks.
}
\vspace{-0.6cm}
\label{fig:main}
}
}
\makeatother

\maketitle

\begin{abstract}
Video spatial reasoning is essential for navigation-oriented perception and long-video question answering, where models must infer spatial relations across long horizons under changing viewpoints.
However, existing multimodal large language models (MLLMs) remain largely semantic-centric, and often fail to reliably aggregate consistent spatial evidence from redundant video observations, leading to inefficient or unstable reasoning.
To address these issues, we propose \textbf{\modelname{}}, a geometry-consistency-aware framework for geometry-sensitive video spatial reasoning that turns spatial consistency into both an evidence organization principle and an explicit post-SFT learning signal.
We build a geometry-consistent memory (GCM) including implicit evidence tokens and explicit geometric cues, and leverage efficient organization strategies to compactly preserve task-related spatial evidence.
Furthermore, we utilize unified consistency self-supervised reinforcement learning (UC-SSRL) after supervised fine-tuning to improve cross-view stability, with answer-, metric-, and topology-consistency rewards.
Extensive experiments on three spatial-reasoning benchmarks, VSI-Bench, OSI-Bench, and MMSI-Video-Bench, show consistent gains, improving the average score by \textbf{12.6} points over the strongest baselines.

\end{abstract}

\section{Introduction}
Video spatial reasoning is a core capability for navigation-oriented perception~\cite{ramakrishnan2021habitat,krantz2020beyond,savva2019habitat,liu2025nav,huang2025mobilevla,wang2026lamp} and video question answering~\cite{wang2025lvbench,chen2025cgbench,wang2026let,zhang2026multigranularity}.
It requires agents and models to infer object locations, scene connectivity, directions, and spatial relations under partial observability or long video contexts.
These demands make video spatial reasoning a practical challenge at the intersection of long-context understanding and physical-space reasoning.

Despite rapid progress in multimodal large language models (MLLMs)\cite{bai2025qwen3,team2023gemini,singh2025openai,zhi2025lscenellm,huang20253d,tang20263d,huang20253dcoca,huang2025dc}, recent spatial benchmarks\cite{yang2025thinking,lin2025mmsi,wu2025indoor} show that reliable spatial reasoning remains a key bottleneck.
Many video-MLLMs are semantics-first: they emphasize recognition and language-conditioned reasoning, while modeling viewpoint geometry and spatial structure only weakly.
Recent works address this limitation through spatial data, training objectives, and reasoning formats, such as SpaceR~\cite{ouyang2025spacer}, ViLaSR~\cite{wu2025reinforcing}, Cambrian-S~\cite{yang2025cambrian}, and RoboBrain~\cite{ji2025robobrain}.
Another direction incorporates explicit spatial cues into MLLMs.
Feed-forward geometry foundation models, such as DUSt3R~\cite{wang2024dust3r}, MASt3R~\cite{leroy2024grounding}, and VGGT~\cite{wang2025vggt}, make pose, depth, and structure signals available from RGB videos.
Built on these geometry priors, Spatial-MLLM~\cite{wu2025spatial}, VLM-3R~\cite{fan2025vlm}, SpaceMind~\cite{zhao2025spacemind}, and GeoThinker~\cite{li2026thinking} fuse geometry-aware tokens into MLLMs.
Other methods exploit explicit 3D representations or maps for stronger grounding and interpretability, including Video-3D LLM~\cite{zheng2025video}, N3D-VLM~\cite{wang2025n3d}, GS-Reasoner~\cite{chen2026reasoning}, and Map2Thought~\cite{gao2026map2thought}.

However, video spatial reasoning still faces two practical challenges: efficiency and robustness under viewpoint changes.
Adjacent frames often contain redundant spatial evidence, so storing or attending to more frames can increase memory and computation without proportional gains, as shown in Fig.~\ref{fig:main}(a).
Moreover, redundant or mismatched evidence may cause inconsistent spatial grounding and unstable answers across viewpoints, as shown in Fig.~\ref{fig:main}(b).
We argue that both challenges can be addressed by exploiting \emph{spatial consistency}: under smooth viewpoint changes of a static scene, correct spatial relations should remain invariant, providing a natural signal for compact evidence organization and stable reasoning.

To this end, we propose \textbf{\modelname{}}, a \emph{geometry-consistency-aware} framework for video spatial reasoning.
\modelname{} uses 3D spatial consistency as both an evidence organization principle and a post-SFT learning objective.
It introduces a \emph{Geometry-Consistent Memory} that stores visual-spatial evidence with explicit pose and depth cues, and organizes them through geometry-guided writing, fusion, and retrieval for compact and robust reasoning.
To further improve viewpoint stability, we propose Unified Consistency Self-Supervised Reinforcement Learning (UC-SSRL), which refines answer, metric, and topological consistency across paired views after supervised fine-tuning anchors answer correctness.
Extensive experiments on \textbf{MMSI-Video-Bench}~\cite{lin2025mmsi}, \textbf{OSI-Bench}~\cite{wu2025indoor}, and \textbf{VSI-Bench}~\cite{yang2025thinking} demonstrate the effectiveness of \modelname{} across spatial construction, relational, and metric reasoning tasks.
Overall, \modelname{} improves the average score by \textbf{12.6} points over the strongest baselines, as illustrated in Fig.~\ref{fig:main}(c).

Our main contributions are as follows:
\begin{itemize}
\item We propose \textbf{\modelname{}}, a geometry-consistency-aware framework for video spatial reasoning that organizes spatial evidence and improves viewpoint stability.
\item We introduce a Geometry-Consistent Dual-Memory with geometry-guided writing, fusion, and retrieval, together with UC-SSRL for post-SFT consistency learning.
\item Extensive experiments and analyses on three spatial-reasoning benchmarks show that \modelname{} improves the average score by \textbf{12.6} points over the strongest baselines.
\end{itemize}



\section{Related Work}
\noindent\textbf{Multimodal large language models.}
MLLMs~\cite{bai2025qwen3,team2023gemini,singh2025openai} have advanced general image and video understanding.
Recent long-video MLLMs~\cite{chen2025longvila} and memory-augmented agents~\cite{wang2024videoagent} improve temporal scalability through longer contexts, retrieval, or external memory.
However, recent spatial benchmarks\cite{yang2025thinking,lin2025mmsi,dongfang2026multimodal,li2025viewspatial,zhang2026flatland} show that reliable spatial reasoning remains challenging.
General-purpose MLLMs are strong in semantic perception and language-conditioned reasoning, but they often lack explicit mechanisms for maintaining spatial consistency across viewpoints and long temporal contexts.
Our work builds on MLLMs, while focusing on geometry-sensitive video spatial reasoning rather than general video understanding.

\noindent\textbf{Spatial-aware MLLMs for spatial reasoning.}
Prior work improves spatial reasoning through 3D cues, spatial supervision, and reasoning-centric training.
Some methods inject explicit 3D information, such as Video-3D LLM~\cite{zheng2025video} and N3D-VLM~\cite{wang2025n3d}.
Others rely on reconstruction- or consistency-based supervision, such as RoSS3D~\cite{wang2025ross3d} and VLM-3R~\cite{fan2025vlm}, or scale spatial data and training, such as Cambrian-S~\cite{yang2025cambrian}.
Reasoning-centric methods further enhance complex spatial reasoning with RL, grounding-aware CoT, or 3D-centric reasoning, including SpatialLadder~\cite{li2025spatialladder}, GS-Reasoner~\cite{chen2026reasoning}, and Think3D~\cite{zhang2026think3d}.
Several methods also design geometry-aware integration modules, such as SpaceMind~\cite{zhao2025spacemind} and GeoThinker~\cite{li2026thinking}.
Despite strong progress, these methods mainly focus on representation design or training-time supervision, with limited control over how redundant video evidence is written, fused, and retrieved.
In contrast, we use geometric consistency as an explicit criterion for managing the full memory lifecycle.

\noindent\textbf{MLLMs with geometry foundation models.}
Geometry foundation models provide strong 3D priors from monocular or multi-view imagery, enabling geometry-aware perception without full 3D supervision.
Representative examples include DUSt3R-style reconstruction~\cite{wang2024dust3r}, feed-forward geometry with VGGT~\cite{wang2025vggt}, permutation-equivariant geometry learning ($\pi^3$)~\cite{wang2026pi}, and depth foundations such as Depth Anything 3~\cite{lin2025depth}.
These models are typically incorporated into MLLMs via feature fusion, distillation, or alignment, as in VG-LLM~\cite{zheng2025learning}, G$^2$VLM~\cite{hu2025g}, 3DRS~\cite{huang20253drs}, Spatial Forcing~\cite{li2025spatial}, and Spatial-MLLM~\cite{wu2025spatial}.
However, geometry priors are mostly used as input features or training signals, while their role in long-video evidence management remains less explored.
We instead treat geometry as a consistency criterion for deciding when to write, how to fuse, and what to retain during hierarchical retrieval.

\begin{figure}[t]
    \centering
    \small
    \includegraphics[width=\linewidth]{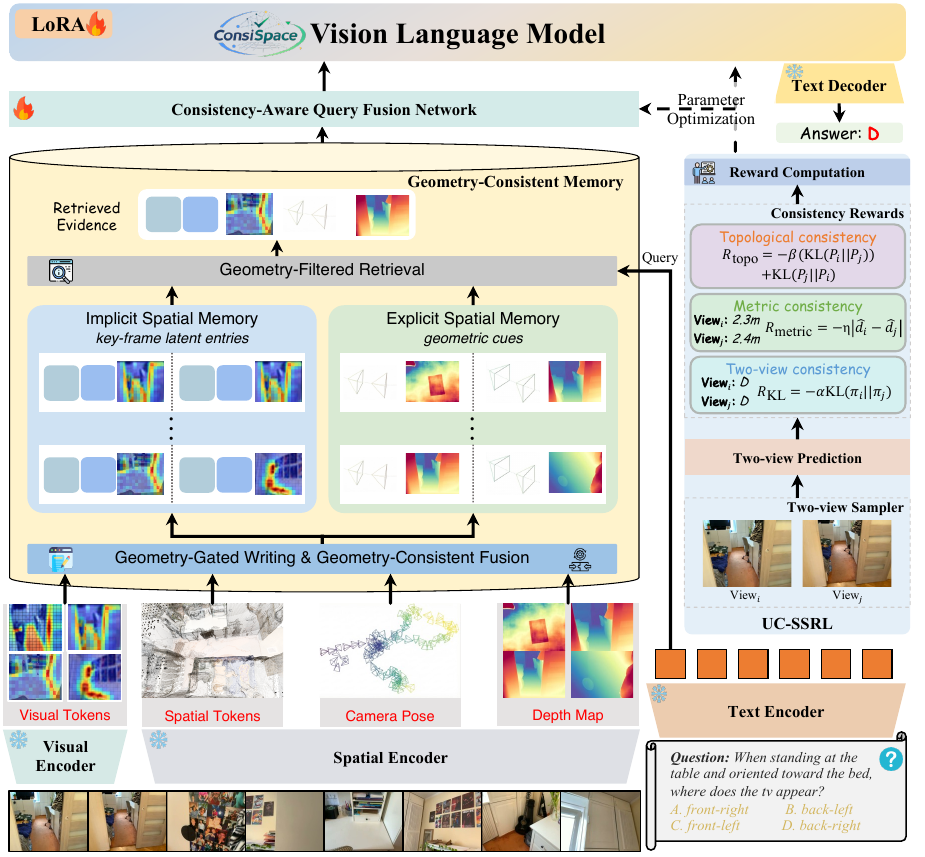}
    \small
    \caption{\textbf{Overview of \modelname{}.}
    \modelname{} organizes visual-spatial tokens and pose-depth cues with a geometry-consistent memory, retrieves query-relevant evidence for answer generation, and uses UC-SSRL to improve viewpoint stability.
    }
    \label{fig:overview}
    \vspace{-0.7cm}
\end{figure}

\section{Method}

\subsection{Overview}
We consider video spatial reasoning, where a model answers a query $q$ given a video $V=\{I_t\}_{t=1}^T$, producing an output $y$.
This setting requires reliable spatial inference over long temporal horizons, while observations are often redundant and spatial evidence can be sparse or viewpoint-dependent.
\modelname{} addresses this problem by using \textbf{geometric consistency} to organize video evidence and stabilize cross-view reasoning.
As illustrated in Fig.~\ref{fig:overview}, \modelname{} consists of dual encoders, a geometry-consistent memory (GCM) with update and read operations, a query-conditioned fusion module, and a video language model for prediction.
During training, we start from the supervised fine-tuned model and apply unified consistency self-supervised reinforcement learning (UC-SSRL) to refine viewpoint stability with self-supervised consistency rewards.

\subsection{Feature Extraction}
\noindent\textbf{Visual tokens.}
Given each frame $I_t$, we extract patch-level visual embeddings using a frozen SigLIP2~\cite{tschannen2025siglip} vision encoder $E_{\text{vis}}$:
\begin{equation}
X_t = E_{\text{vis}}(I_t), \qquad X_t \in \mathbb{R}^{N \times d},
\end{equation}
where $N$ denotes the number of visual tokens and $d$ is the embedding dimension.
These tokens provide semantic evidence for memory construction and retrieval.

\noindent\textbf{Spatial tokens and geometric cues.}
We extract geometry-aware representations with a frozen VGGT~\cite{wang2025vggt} geometry encoder $E_{\text{spa}}$.
Given a frame $I_t$ optionally conditioned on preceding frames $I_{<t}$, VGGT produces spatial tokens and explicit geometric cues:
\begin{equation}
S_t, G_t = E_{\text{spa}}(I_t\,;\, I_{<t}),
\end{equation}
where $S_t$ denotes geometry-aware spatial tokens, and $G_t$ includes camera pose $T_t$ and per-frame depth map $D_t$.
These cues provide geometric consistency signals for memory writing, fusion, and retrieval.

\begin{figure}[t]
    \centering
    \includegraphics[width=\linewidth]{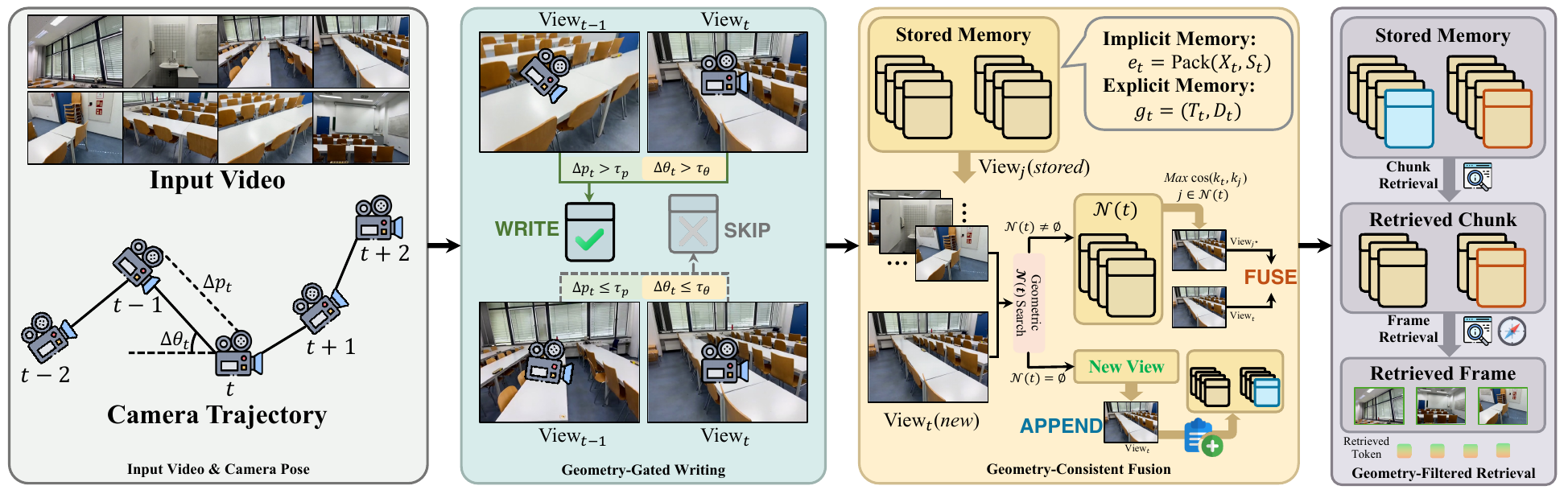}
    \small
    \caption{\textbf{Geometry-Consistent Memory (GCM) evidence lifecycle.}
    GCM uses geometry-gated writing, geometry-consistent fusion, and geometry-filtered hierarchical retrieval to compact redundant views and retrieve query-relevant evidence.
    }
    \label{fig:gcm_life}
    \vspace{-0.7cm}
\end{figure}

\subsection{Geometry-Consistent Memory}
\modelname{} maintains a \emph{Geometry-Consistent Memory} $\mathcal{M}$ with two complementary banks (Fig.~\ref{fig:overview}).
The Implicit Spatial Memory $\mathcal{M}_{\text{imp}}$ stores per-frame evidence tokens, including visual semantics and spatial tokens, for query-relevant retrieval.
The Explicit Spatial Memory $\mathcal{M}_{\text{exp}}$ stores camera poses and depth cues, which provide geometry-based consistency signals for memory control.
This separation decouples evidence representation from geometry-guided operations: $\mathcal{M}_{\text{imp}}$ supports evidence retrieval, while $\mathcal{M}_{\text{exp}}$ guides geometry-gated writing, geometry-consistent fusion, and geometry-filtered retrieval (Fig.~\ref{fig:gcm_life}).

\noindent\textbf{Memory entries.}
For each frame $t$, we construct an implicit evidence entry by packing visual tokens $X_t$ and spatial tokens $S_t$:
\begin{equation}
e_t=\textsc{Pack}(X_t, S_t), \qquad g_t=(T_t, D_t),
\end{equation}
where $T_t$ is the camera pose and $D_t$ is the per-frame depth map predicted by the spatial encoder.
$\textsc{Pack}(\cdot)$ concatenates and projects visual and spatial tokens into a fixed-length evidence representation.
We store $e_t$ in $\mathcal{M}_{\text{imp}}$ and the aligned geometric cues $g_t$ in $\mathcal{M}_{\text{exp}}$ with the same timestamp.

\noindent\textbf{Geometry-gated writing.}
Adjacent frames are often redundant, so writing every frame to memory is inefficient.
We gate memory writes using geometric changes predicted by the geometry encoder.
Let $T_t$ be the camera pose, and let $(p_t,v_t)$ denote the camera position and viewing direction derived from $T_t$.
We compute inter-frame motion and viewpoint changes:
\begin{equation}
\Delta p_t=\|p_t-p_{t-1}\|_2,\qquad
\Delta \theta_t=\angle(v_t,v_{t-1}),
\end{equation}
and write an entry only when new spatial evidence is likely to appear:
\begin{equation}
\textsc{write}(t)=\mathbb{I}\left[\Delta p_t>\tau_p\ \lor\ \Delta \theta_t>\tau_\theta\right].
\end{equation}
This geometry-gated policy suppresses redundant writes under smooth viewpoint changes while preserving informative transitions.

\noindent\textbf{Geometry-consistent fusion.}
When \textsc{write}$(t)=1$, we further consolidate entries that are geometrically consistent.
We define a geometric neighborhood for entry $t$:
\begin{equation}
\mathcal{N}(t)=\left\{j:\ \|p_t-p_j\|<r,\ \angle(v_t,v_j)<\theta \right\}.
\end{equation}
If $\mathcal{N}(t)\neq \emptyset$, we select a fusion target based on semantic similarity in the evidence space:
\begin{equation}
j^\star=\arg\max_{j\in\mathcal{N}(t)} \cos(k_t,k_j),\qquad k_t = W_k e_t,
\end{equation}
where $k_t$ is a key projected from the evidence entry $e_t$.
We then fuse $e_t$ into $e_{j^\star}$; otherwise, we append $(e_t,g_t)$ as a new entry.
This step reduces repeated evidence while maintaining a geometry-consistent scene memory.

\noindent\textbf{Geometry-filtered retrieval.}
Given a query $q$, we obtain a text embedding $q_e$ via a text encoder $E_{\text{text}}$ and perform coarse-to-fine retrieval.
We first retrieve candidate chunks based on semantic relevance:
\begin{equation}
s_c(j)=\cos(q_e,k_j^c), \qquad
C_{\text{topk}}=\textsc{TopK}\!\left(\{s_c(j)\},K_c\right),
\end{equation}
which prioritizes recall at a low cost.

Within candidate chunks, we refine frame-level selection by combining semantic relevance with a lightweight geometry term:
\begin{equation}
s_f(i)=\alpha \cos(q_e,k_i^f)+\beta \cos(Q_{\text{topo}}, e_i^{\text{dir}}),
\end{equation}
where $Q_{\text{topo}}$ is a direction-oriented embedding derived from $q$, and $e_i^{\text{dir}}$ encodes the viewing direction derived from $T_i$.
We select top-$K_f$ frames $F_{\text{topk}}$ according to $s_f(i)$ and gather the corresponding evidence.

In addition to the retrieved evidence, we construct lightweight summary tokens to provide global topological and metric context.
Given $q_e$, we derive two query vectors:
\begin{equation}
Q_{\text{topo}} = W_{\text{topo}} q_e,\qquad
Q_{\text{metric}} = W_{\text{metric}} q_e,
\end{equation}
where $Q_{\text{topo}}$ emphasizes topological cues such as relative direction or connectivity, and $Q_{\text{metric}}$ emphasizes metric cues such as distance, size, and ordering.
We compute summary tokens via query-conditioned aggregation:
\begin{equation}
z_{\text{topo}}=\mathrm{CA}(Q_{\text{topo}}, \mathcal{B}_{\text{topo}}),\qquad
z_{\text{metric}}=\mathrm{CA}(Q_{\text{metric}}, \mathcal{B}_{\text{metric}}),
\end{equation}
where $\mathrm{CA}(\cdot)$ denotes a lightweight query-conditioned aggregation operator.
$\mathcal{B}_{\text{topo}}$ contains values from $\mathcal{M}_{\text{imp}}$, and $\mathcal{B}_{\text{metric}}$ contains embedded geometric cues from $\mathcal{M}_{\text{exp}}$.

Finally, we form the context for the video language model by concatenating the query token, summary tokens, retrieved evidence tokens, and aligned geometric cues:
\begin{equation}
Z=\mathrm{Concat}\big([\mathrm{TOK}(q)],\, z_{\text{topo}}, z_{\text{metric}},\, \{z_i\}_{i\in F_{\text{topk}}},\, \{g_i\}_{i\in F_{\text{topk}}}\big),
\end{equation}
where $\{z_i\}$ are evidence tokens from $\mathcal{M}_{\text{imp}}$, and $\{g_i\}$ are aligned geometric cues from $\mathcal{M}_{\text{exp}}$.
In practice, geometry filtering can be implemented by down-weighting inconsistent entries during scoring or by masking them before concatenation.
The resulting sequence $Z$ is projected into the language model embedding space and fed into the video language model for prediction.

\subsection{Unified Consistency Self-Supervised Reinforcement Learning}
While geometry-consistent memory organizes evidence efficiently, the model may still produce different answers when the same question is observed from different temporal windows or retrieved evidence sets.
Such inconsistency indicates brittle spatial grounding under viewpoint changes.
We therefore optimize \emph{cross-view geometric consistency} after supervised fine-tuning, where SFT provides task-aligned initialization and UC-SSRL further refines viewpoint stability.
Different from Spatial-SSRL~\cite{liu2025spatial}, which relies on intrinsic pretext tasks, our objective derives self-supervised rewards from multi-view consistency in video spatial reasoning: if the underlying 3D relations are stable, predictions from different views should agree.

\noindent\textbf{Two-view sampler.}
Given a training sample $(V,q)$, we construct two observations for the same question, e.g., two temporal windows, two retrieval contexts, or two evidence sets:
\begin{equation}
o_i = \mathcal{S}(V,q;\xi_i), \qquad o_j = \mathcal{S}(V,q;\xi_j),
\end{equation}
where $\mathcal{S}(\cdot)$ denotes the two-view sampler and $\xi_i,\xi_j$ are stochastic sampling choices.
Feeding $(o_i,q)$ and $(o_j,q)$ into the model yields two predicted answer distributions $\pi_i,\pi_j$.

\noindent\textbf{Unified consistency rewards.}
We design three complementary self-supervised rewards to refine answer-level, metric, and relational stability across paired views.

\emph{Two-view answer consistency.}
We encourage agreement between the two predicted answer distributions by penalizing their divergence:
\begin{equation}
R_{\mathrm{KL}} = -\alpha\,\mathrm{KL}(\pi_i \,\|\, \pi_j),
\end{equation}
which stabilizes answers across different views.

\emph{Metric consistency.}
For metric-related questions, such as distance and size, let $\hat d_i$ and $\hat d_j$ denote the distance estimates inferred from the two views.
We enforce consistency via
\begin{equation}
R_{\mathrm{metric}} = -\eta\,|\hat d_i - \hat d_j|.
\end{equation}

\emph{Topological consistency.}
For relational reasoning, let $P_i$ and $P_j$ denote normalized relational distributions derived from the two views.
We encourage symmetric agreement:
\begin{equation}
R_{\mathrm{topo}} = -\beta\left(\mathrm{KL}(P_i \,\|\, P_j)+\mathrm{KL}(P_j \,\|\, P_i)\right).
\end{equation}

We combine them into a unified reward:
\begin{equation}
R = R_{\mathrm{KL}} + R_{\mathrm{metric}} + R_{\mathrm{topo}},
\end{equation}
and optimize the model with a policy-gradient style objective.
UC-SSRL is initialized from the SFT model and refines consistency without additional human annotations; in our implementation, only LoRA parameters are updated (Fig.~\ref{fig:overview}), making it lightweight and complementary to memory-based evidence organization.

\begin{table}[t]
\centering
\small
\caption{\textbf{VSI-Bench results.}
``Avg.’’ denotes the overall mean.
\textbf{Bold} and \underline{underline} denote the best and second-best results.
}
\vspace{-0.35cm}
\resizebox{0.9\linewidth}{!}{
\begin{tabular}{l|c|cccc cccc}
\toprule
& & \rotatebox{60}{Obj. Count} & \rotatebox{60}{Abs. Dist.} & \rotatebox{60}{Obj. Size} & \rotatebox{60}{Room Size}
& \rotatebox{60}{Rel. Dist.} & \rotatebox{60}{Rel. Dir.} & \rotatebox{60}{Route Plan} & \rotatebox{60}{Appr. Order} \\
\textbf{Method} & \textbf{Avg.}
& \multicolumn{4}{c}{\cellcolor{orange!10}\textbf{Numerical (MRA)}}
& \multicolumn{4}{c}{\cellcolor{yellow!10}\textbf{Multiple-Choice (Acc.)}} \\
\hline
\rowcolor{green!10}
\multicolumn{1}{l|}{\textbf{\textit{Proprietary Models (API)}}} & & & & & & & & & \\
GPT-4o            & 34.0 & 46.2 &  5.3 & 43.8 & 38.2 & 37.0 & 41.3 & 31.5 & 28.5 \\
Gemini-1.5 Flash  & 42.1 & 49.8 & 30.8 & 53.5 & 54.4 & 37.7 & 41.0 & 31.5 & 37.8 \\
Gemini-1.5 Pro    & 45.4 & 56.2 & 30.9 & 64.1 & 43.6 & 51.3 & 46.3 & 36.0 & 34.6 \\
\hline
\rowcolor{green!10}
\multicolumn{1}{l|}{\textbf{\textit{Open-source Models}}} & & & & & & & & & \\
LLaVA-Video-7B                     & 35.6 & 48.5 & 14.0 & 47.8 & 24.2 & 43.5 & 42.4 & 34.0 & 30.6 \\
LLaVA-Video-72B                    & 40.9 & 48.9 & 22.8 & 57.4 & 35.3 & 42.4 & 36.7 & 35.0 & 48.6 \\
LLaVA-OneVision-7B                 & 32.4 & 47.7 & 20.2 & 47.4 & 12.3 & 42.5 & 35.2 & 29.4 & 24.4 \\
Qwen2.5-VL-7B                      & 33.0 & 40.9 & 14.8 & 43.4 & 10.7 & 38.6 & 38.5 & 33.0 & 29.8 \\
LLaVA-OneVision-72B                & 40.2 & 43.5 & 23.9 & 57.6 & 37.5 & 42.5 & 39.9 & 32.5 & 44.6 \\
Qwen3-VL-8B-Instruct               & 57.4 & 66.7 & 38.9 & 74.1 & 53.5 & 59.0 & 65.4 & 32.0 & 69.4 \\  
\hline
\rowcolor{green!10}
\multicolumn{1}{l|}{\textbf{\textit{Specialized Models}}} & & & & & & & & & \\
Spacer~\cite{ouyang2025spacer}     & 45.5 & 57.8 & 28.2 & 59.9 & 47.1 & 40.1 & 45.4 & 33.5 & 52.1 \\
ViLaSR~\cite{wu2025reinforcing}    & 45.4 & 63.5 & 34.4 & 60.6 & 30.9 & 48.9 & 45.2 & 30.4 & 49.2 \\
Spatial-MLLM~\cite{wu2025spatial}  & 48.4 & 65.3 & 34.8 & 63.1 & 45.1 & 41.3 & 46.2 & 33.5 & 46.3 \\
GeoThinker~\cite{li2026thinking}   & 50.5 & 69.5 & 38.5 & 57.9 & 62.2 & 45.2 & 46.2 & 31.4 & 52.6 \\
VLM-3R~\cite{fan2025vlm}           & 60.9 & 70.2 & 49.4 & 69.2 & 67.1 & 65.4 & 80.5 & 45.4 & 40.1 \\
Map2Thought~\cite{gao2026map2thought} & 61.0 & 70.8 & 55.0 & 70.1 & 69.4 & 56.9 & 69.8 & 38.1 & 57.4 \\
Cambrian-S-7B~\cite{yang2025cambrian}& 67.5 & 73.2 & 50.5 & 74.9 & 72.2 & \underline{71.1} & 76.2 & 41.8 & \textbf{80.1} \\
SpaceMind~\cite{zhao2025spacemind} & 69.6 & \underline{73.3} & 61.4 & 77.3 & \underline{74.2} & 67.2 & \underline{88.4} & 44.3 & 70.6 \\
\hline
\textbf{\modelname{}}\emph{$_{SFT}$}          & \underline{71.2} & 71.0 & \underline{73.8} & \underline{77.7} & 71.7 & 68.5 & 83.6 & \underline{49.8} & 73.8 \\
\textbf{\modelname{}}\emph{$_{UC-SSRL}$}         & \textbf{76.6} & \textbf{77.9} & \textbf{78.9} & \textbf{82.2} & \textbf{76.3} & \textbf{72.1} & \textbf{90.1} & \textbf{54.7} & \underline{79.9} \\
\bottomrule
\end{tabular}
}
\label{tab:vsi_bench_tiny}
\vspace{-0.8cm}
\end{table}

\section{Experiments}

\subsection{Evaluation Benchmarks}

\noindent\textbf{VSI-Bench.}
VSI-Bench~\cite{yang2025thinking} evaluates visual-spatial intelligence from indoor-scene videos, requiring models to infer spatial relations and measurements from continuous observations.
It contains over 5,000 question-answer pairs from 288 real-world indoor videos, covering configurational, measurement-estimation, and spatiotemporal tasks.
Following its protocol, we report \textbf{Accuracy} for multiple-choice answer (MCA) tasks and \textbf{Mean Relative Accuracy (MRA)} for numerical-answer (NA) tasks.
For prediction $\hat{y}$ and ground truth $y$, MRA is computed over tolerance thresholds $\mathcal{C}=\{0.50,0.55,\ldots,0.95\}$:
\begin{equation}
\mathrm{MRA}=\frac{1}{|\mathcal{C}|}\sum_{\theta\in\mathcal{C}}\mathbb{I}\left(\frac{|\hat{y}-y|}{y} < 1-\theta\right).
\end{equation}

\noindent\textbf{OSI-Bench.}
OSI-Bench~\cite{wu2025indoor} targets open-world spatial reasoning from pedestrian-perspective videos with metrically precise 3D ground truth from synchronized multi-sensor capture.
It includes both qualitative relational reasoning and quantitative metric or kinematic understanding, making it suitable for testing whether spatial reasoning gains transfer beyond indoor benchmarks.
Following its protocol, we report \textbf{Accuracy} for MCA tasks and \textbf{MRA} for NA tasks, where MRA checks whether the relative error falls below a set of tolerances with a small-value floor for near-zero targets.

\noindent\textbf{MMSI-Video-Bench.}
MMSI-Video-Bench~\cite{lin2025mmsi} evaluates video-based spatial intelligence with fully human-annotated questions from diverse video sources.
It contains 1,106 questions over 1,278 clips and covers Perception, Planning, Prediction, and Cross-Video Reasoning.
Following the official protocol, we report exact-match accuracy under two input settings:
\textbf{Uniform-50}, which feeds exactly 50 uniformly sampled frames per clip, and
\textbf{Sufficient-Coverage}, which uses the full set of frames used during human annotation.
These settings evaluate spatial reasoning under a fixed frame budget and under evidence-rich inputs, respectively.

\begin{table}[t]
\centering
\small
\caption{\textbf{OSI-Bench results.}
``Avg.’’ denotes the overall mean.
\textbf{Bold} and \underline{underline} denote the best and second-best results.
}
\vspace{-0.35cm}
\label{tab:osi_bench}
\resizebox{0.9\linewidth}{!}{
\begin{tabular}{l|c|ccccccccc}
\toprule
& & \rotatebox{60}{Rel. Dis.} & \rotatebox{60}{Rel. Dir.} & \rotatebox{60}{Qual. EM} & \rotatebox{60}{Obj. Loc.}
& \rotatebox{60}{Abs. Dis.} & \rotatebox{60}{Depth Count} & \rotatebox{60}{Abs. Displ.}
& \rotatebox{60}{Abs. Speed} & \rotatebox{60}{Quan. EM} \\
\textbf{Method} & \textbf{Avg.}
& \multicolumn{3}{c}{\cellcolor{green!20}\textbf{Relational $(\mathcal{MCA})$}}
& \multicolumn{3}{c}{\cellcolor{purple!20}\textbf{Static Metric $(\mathcal{NA})$}}
& \multicolumn{3}{c}{\cellcolor{pink!20}\textbf{Dynamic Metric $(\mathcal{NA})$}} \\
\hline
\rowcolor{green!10}
\multicolumn{1}{l|}{\textbf{\textit{Proprietary Models (API)}}} & & & & & & & & & & \\
GPT-5                  & 29.7 & 34.4 & 33.1 & 49.5 & 32.5 & 23.7 & 20.9 & 10.5 & 33.8 & 30.6 \\
Gemini-2.5-Pro         & 37.2 & 50.0 & 28.1 & 52.5 & 37.4 & 28.1 & 37.9 & 26.8 & 31.1 & 40.8 \\
\hline
\rowcolor{green!10}
\multicolumn{1}{l|}{\textbf{\textit{Open-source Models}}} & & & & & & & & & & \\
LLaVA-Video-Qwen2-7B   & 22.9 & 37.1 & 31.2 & 40.9 & 17.6 & 22.1 & 18.2 & 17.5 & 19.0 &  5.7 \\
LLaVA-OneVision-7B     & 25.7 & 35.1 & 32.7 & 40.8 & 16.1 & 25.5 & 25.6 & 16.8 & 28.3 & 13.4 \\
Qwen3-VL-8B-Instruct   & 31.2 & 38.3 & 31.2 & 49.3 & 21.0 & 15.1 & 33.3 & 21.3 & 34.3 & 37.8 \\
\hline
\rowcolor{green!10}
\multicolumn{1}{l|}{\textbf{\textit{Specialized Models}}} & & & & & & & & & & \\
Spacer~\cite{ouyang2025spacer}                       & 30.2 & 35.0 & 24.6 & 39.0 & 36.7 & 14.0 & 36.4 & 24.0 & 28.4 & 34.1 \\
ViLaSR~\cite{wu2025reinforcing}                      & 30.4 & 42.6 & 24.5 & 39.0 & 40.3 & 17.1 & 36.8 & 15.7 & 25.8 & 32.2 \\
Spatial-MLLM~\cite{wu2025spatial}                    & 31.3 & 36.0 & 25.0 & 41.5 & 41.5 & 17.3 & 38.3 & 23.0 & 28.4 & 30.3 \\
VLM-3R~\cite{fan2025vlm}                             & 40.3 & 57.0 & 43.6 & 52.3 & \underline{44.6} & 24.6 & 42.0 & 34.2 & 38.5 & 26.2 \\ \hline
\textbf{\modelname{}}\emph{$_{SFT}$}           & \underline{45.8} & \underline{58.1} & \underline{44.2} & \underline{60.1} & 44.4 & \underline{35.5} & \underline{46.0} & \underline{34.7} & \underline{41.7} & \underline{47.7} \\
\textbf{\modelname{}}\emph{$_{UC-SSRL}$}          & \textbf{53.0} & \textbf{66.5} & \textbf{51.3} & \textbf{68.7} & \textbf{51.6} & \textbf{41.7} & \textbf{52.7} & \textbf{41.8} & \textbf{48.5} & \textbf{54.4} \\
\bottomrule
\end{tabular}
}
\vspace{-0.8cm}
\end{table}
\subsection{Datasets}
\noindent\textbf{Supervised fine-tuning.}
For SFT, we fine-tune the base video language model on a mixture of indoor and outdoor video spatial reasoning data.
For indoor spatial reasoning, we use VSI-590K~\cite{yang2025cambrian} as the primary instruction-tuning source.
To improve outdoor coverage, we additionally construct nuScenes-10K, an outdoor synthetic set of 10K QA instances from nuScenes~\cite{nuscenes2019}, using an automated and programmatically verifiable pipeline based on ego poses and 3D annotations.
We apply geometry-validity filtering, answer parseability checking, consistency checking, MLLM verification, and de-duplication to 22,518 generated candidates, yielding 10,000 QA pairs.
nuScenes-10K has no shared videos or annotations with OSI-Bench, except for generic object-category overlap.
Manual verification on 1,000 randomly sampled QA pairs shows 91.6\% correctness.
Additional details and verification statistics are provided in \textit{Supp. Mat.}~\ref{sec:supp_data_synthesis}.
SFT is performed with standard maximum-likelihood training to map the concatenated context to the final answer.

\noindent\textbf{Unified consistency SSRL.}
UC-SSRL does not require additional human annotations beyond the SFT data.
We reuse the same training pool and apply a two-view sampler to construct paired observations for each $(V,q)$, such as two temporal windows or two retrieval contexts.
The model is then optimized with answer, metric, and topological consistency rewards across paired views.
Unless otherwise stated, only lightweight LoRA parameters are updated during UC-SSRL, making it complementary to SFT and efficient to run.

\subsection{Implementation Details}
\noindent\textbf{Architecture.}
\modelname{} is built on \textbf{Qwen3-VL-8B-Instruct}~\cite{bai2025qwen3} with LoRA adaptation and Geometry-Consistent Memory.
We use frozen SigLIP2~\cite{tschannen2025siglip} as the visual encoder and frozen VGGT~\cite{wang2025vggt} as the geometry encoder.
The vision tower, aligner, and geometry encoder remain frozen, and only LoRA parameters are updated during both SFT and UC-SSRL.

\noindent\textbf{Memory and retrieval.}
We use a two-bank Geometry-Consistent Memory with $M{=}64$ memory tokens and feature dimension 512.
The memory is organized into 4 chunks with an 8-head resampler.
Hierarchical retrieval uses top-$K{=}8$ at both chunk and frame levels, with 8 value tokens per chunk and 4 value tokens per frame.
Detailed memory injection, gating, retrieval, and backend settings are provided in \textit{Supp. Mat.}~\ref{sec:supp_training_details}.

\noindent\textbf{Training details.}
We first fine-tune on VSI-590K and nuScenes-10K with standard maximum-likelihood training.
We use LoRA with rank 16 and $\alpha{=}32$, bfloat16 precision, and DeepSpeed ZeRO-3.
The trainable parameters account for only a small fraction of the full model parameters, since the base model and encoders are frozen.
We train for 200 steps with learning rate $1\times10^{-4}$, per-device batch size 1, gradient accumulation 2, and maximum sequence length 5,120.
We then initialize UC-SSRL from the SFT checkpoint, keep the same optimization setting for controlled comparison, and set the overall UC-SSRL weight to $10^{-2}$.
All experiments are conducted on 8$\times$A100 80GB GPUs.
Additional optimizer, LoRA, and runtime details are provided in \textit{Supp. Mat.}~\ref{sec:supp_training_details}

\begin{table}[t]
\centering
\small
\caption{\textbf{MMSI-Video-Bench results.}
``Avg.’’ denotes the overall mean.
Attr., Inst., Cam., Scen., and Inter. denote Attribute, Instance, Camera, Scene, and Interaction, respectively.
MU., and MV. represent Memory Update, and Multi-View Integration, respectively.
\textbf{Bold} and \underline{underline} denote the best and second-best results within each setting.
}
\vspace{-0.35cm}
\resizebox{\linewidth}{!}{%
\begin{tabular}{l|ccccccccccccc|c}
\toprule
& \rotatebox{60}{Attr.} & \rotatebox{60}{Inst.-Inst.} & \rotatebox{60}{Inst.-Scen.} & \rotatebox{60}{Scen.-Scen.} & \rotatebox{60}{Cam.-Inst.} & \rotatebox{60}{Cam.-Scen.} &
\rotatebox{60}{Cam.} & \rotatebox{60}{Inst.} & \rotatebox{60}{Inter.} &
\rotatebox{60}{MU.} & \rotatebox{60}{MV.} & 
\rotatebox{60}{-} & \rotatebox{60}{-} & \\
\textbf{Method} & \multicolumn{6}{c}{\cellcolor{cyan!8}\textbf{Spatial Construction}} &
\multicolumn{3}{c}{\cellcolor{yellow!30}\textbf{Motion Understanding}} & 
\multicolumn{2}{c}{\cellcolor{green!15}\textbf{Cross-Video}} & \cellcolor{orange!20}\makecell{\textbf{Plan.}} & \cellcolor{purple!25}\makecell{\textbf{Pred.}} & \textbf{Avg.}\\
\hline
\rowcolor{green!10}
\multicolumn{1}{l|}{\textbf{\textit{(Sufficient-Coverage) Proprietary}}} & & & & & & & & & & & & & &\\
GPT-4o                    & 22.9 & 30.3 & 35.1 & 26.1 & 26.6 & 27.5 & 24.7 & 28.9 & 24.7 & 34.3 & 25.7 & 29.8 & 26.8 & 28.1 \\
Gemini 2.5 Flash-Thinking & 43.4 & 38.2 & 41.6 & 24.6 & 34.2 & 36.2 & 44.1 & 34.4 & 35.8 & 36.3 & 38.6 & 36.3 & 31.7 & 36.7 \\
\hline
\rowcolor{green!10}
\multicolumn{1}{l|}{\textbf{\textit{(Sufficient-Coverage) Open-source}}} & & & & & & & & & & & & & &\\
Qwen2.5-VL-7B              & 31.3 & 21.1 & 19.5 & 30.4 & 19.0 & 31.2 & 35.5 & 36.7 & 29.6 & 27.4 & 30.0 & 26.6 & 35.4 & 28.8 \\
Qwen3-VL-8B                & 27.8 & 31.5 & 33.3 & 26.9 & 23.1 & 22.4 & 27.3 & 29.6 & 41.0 & 28.4 & 29.9 & 26.5 & 32.0 & 29.1 \\
\hline
\rowcolor{green!10}
\multicolumn{1}{l|}{\textbf{\textit{(Sufficient-Coverage) Specialized Models}}} & & & & & & & & & & & & & &\\
Spacer~\cite{ouyang2025spacer}             & 35.3 & 29.6 & 39.2 & 27.9 & 35.1 & 27.8 & 36.4 & 28.2 & 33.0 & 35.9 & 30.8 & 28.0 & 27.3 & 31.8 \\
ViLaSR~\cite{wu2025reinforcing}            & 35.2 & 29.5 & 39.1 & 27.8 & 35.0 & 27.7 & 36.3 & 28.1 & 32.9 & 35.8 & 30.7 & 27.9 & 27.2 & 31.7 \\
Spatial-MLLM~\cite{wu2025spatial}          & 37.4 & 31.7 & 41.3 & 30.0 & 37.2 & 29.9 & 38.5 & 30.3 & 35.1 & 38.0 & 32.9 & 30.1 & 29.4 & 33.9 \\
VLM-3R~\cite{fan2025vlm}                   & 46.1 & 40.4 & 50.0 & 38.7 & 45.9 & 38.6 & 47.2 & 39.0 & 43.8 & 46.7 & 41.6 & 38.8 & 38.1 & 42.6 \\
\hline
\textbf{\modelname{}}\emph{$_{SFT}$}        & \underline{51.8} & \underline{46.3} & \underline{55.6} & \underline{44.6} & \underline{51.6} & \underline{44.5} & \underline{52.9} & \underline{44.9} & \underline{49.6} & \underline{52.4} & \underline{47.4} & \underline{44.7} & \underline{44.0} & \underline{48.4} \\
\textbf{\modelname{}}\emph{$_{UC-SSRL}$}        & \textbf{61.4} & \textbf{55.2} & \textbf{65.7} & \textbf{53.3} & \textbf{61.1} & \textbf{53.7} & \textbf{62.2} & \textbf{54.0} & \textbf{58.8} & \textbf{62.0} & \textbf{56.8} & \textbf{53.4} & \textbf{52.9} & \textbf{57.5} \\

\hline \hline

\rowcolor{green!10}
\multicolumn{1}{l|}{\textbf{\textit{(Uniform-50) Proprietary}}} & & & & & & & & & & & & & &\\
Gemini 2.5 Flash-Thinking & 43.4 & 25.0 & 42.9 & 23.2 & 32.9 & 47.5 & 38.7 & 28.9 & 38.3 & 31.4 & 28.6 & 40.3 & 31.7 & 35.2 \\
GPT-5                     & 44.6 & 38.2 & 44.2 & 39.1 & 40.5 & 41.2 & 37.6 & 28.9 & 42.0 & 33.3 & 28.6 & 34.7 & 28.1 & 36.8 \\
Gemini 3 Pro              & 44.6 & 39.5 & 44.2 & 33.3 & 29.1 & 43.8 & 35.5 & 40.0 & 38.3 & 34.3 & 37.1 & 38.7 & 35.4 & 38.0 \\
\hline

\rowcolor{green!10}
\multicolumn{1}{l|}{\textbf{\textit{(Uniform-50) Open-source}}} & & & & & & & & & & & & & &\\
LLaVA-Video-7B            & 27.7 & 28.9 & 28.6 & 30.4 & 19.0 & 25.0 & 31.2 & 35.6 & 30.9 & 36.3 & 20.0 & 21.0 & 35.4 & 28.5 \\
Qwen2.5-VL-7B              & 26.5 & 25.0 & 29.9 & 34.8 & 20.2 & 37.5 & 33.3 & 31.1 & 34.6 & 24.5 & 22.9 & 34.7 & 28.1 & 29.7 \\
Qwen3-VL-8B                & 33.7 & 25.0 & 24.7 & 26.1 & 24.1 & 30.0 & 35.5 & 22.2 & 29.6 & 31.4 & 20.0 & 25.0 & 29.3 & 27.6 \\
\hline
\rowcolor{green!10}
\multicolumn{1}{l|}{\textbf{\textit{(Uniform-50) Specialized Models}}} & & & & & & & & & & & & & &\\
Spacer~\cite{ouyang2025spacer}            & 36.4 & 31.1 & 36.3 & 30.9 & 32.3 & 39.5 & 30.5 & 32.2 & 34.0 & 28.1 & 29.2 & 32.1 & 27.5 & 32.3 \\
ViLaSR~\cite{wu2025reinforcing}           & 36.3 & 31.0 & 36.2 & 30.8 & 32.2 & 39.4 & 30.4 & 32.1 & 33.9 & 28.0 & 29.1 & 32.0 & 27.4 & 32.2 \\
Spatial-MLLM~\cite{wu2025spatial}         & 38.5 & 33.2 & 38.4 & 33.0 & 34.4 & 41.6 & 32.6 & 34.3 & 36.1 & 30.2 & 31.3 & 34.2 & 29.6 & 34.4 \\
VLM-3R~\cite{fan2025vlm}                  & 47.2 & 41.9 & 47.1 & 41.7 & 43.1 & 50.3 & 41.3 & 43.0 & 44.8 & 38.9 & 40.0 & 42.9 & 38.3 & 43.1 \\
\hline
\textbf{\modelname{}}\emph{$_{SFT}$}       & \underline{52.9} & \underline{47.7} & \underline{52.8} & \underline{47.5} & \underline{48.9} & \underline{55.9} & \underline{47.1} & \underline{48.8} & \underline{50.5} & \underline{44.8} & \underline{45.9} & \underline{48.7} & \underline{44.2} & \underline{48.9} \\
\textbf{\modelname{}}\emph{$_{UC-SSRL}$}       & \textbf{62.6} & \textbf{56.7} & \textbf{62.5} & \textbf{56.8} & \textbf{58.0} & \textbf{65.8} & \textbf{56.2} & \textbf{58.2} & \textbf{59.9} & \textbf{54.0} & \textbf{55.0} & \textbf{58.1} & \textbf{53.2} & \textbf{58.1} \\

\bottomrule
\end{tabular}
}
\vspace{-0.8cm}
\label{tab:mmsi_benchmark_results}
\end{table}

\subsection{Main Results}

\noindent\textbf{VSI-Bench.}
As shown in Tab.~\ref{tab:vsi_bench_tiny}, \modelname{} achieves 76.6 Avg. with UC-SSRL, outperforming SpaceMind (69.6 Avg.) by +7.0 points and Qwen3-VL-8B-Instruct (57.4 Avg.) by +19.2 points.
The gains cover both numerical and multiple-choice tasks, showing improved measurement and relational reasoning.

\noindent\textbf{OSI-Bench.}
Tab.~\ref{tab:osi_bench} shows that \modelname{} reaches 53.0 Avg. with UC-SSRL on OSI-Bench.
It improves over VLM-3R (40.3 Avg.) by +12.7 points and Qwen3-VL-8B-Instruct (31.2 Avg.) by +21.8 points.
UC-SSRL further improves over SFT by +7.2 points, especially on metric-heavy categories, suggesting better stability under viewpoint changes.


\noindent\textbf{MMSI-Video-Bench.}
As shown in Tab.~\ref{tab:mmsi_benchmark_results}, \modelname{} achieves 57.5 Avg. under \textit{Sufficient-Coverage} and 58.1 Avg. under \textit{Uniform-50}, outperforming VLM-3R by +14.9 and +15.0 points, respectively.
UC-SSRL consistently improves over SFT in both settings, indicating that consistency learning benefits both evidence-rich and fixed-budget inputs.

\begin{table}[t]
\centering
\small
\caption{\textbf{End-to-end efficiency and long-video scaling.}
All results are measured on the same A100 80GB GPU under different input budgets.
}
\label{tab:efficiency}
\vspace{-0.35cm}
\resizebox{0.8\linewidth}{!}{
\begin{tabular}{lcccccccc}
\toprule
\textbf{Method} & \textbf{Dur.} & \textbf{Frames} & \textbf{Mem. (GB)}$\downarrow$ & \textbf{Inf. (s)}$\downarrow$
& \textbf{VSI}$\uparrow$ & \textbf{OSI}$\uparrow$ & \textbf{MMSI-SC}$\uparrow$ & \textbf{MMSI-U50}$\uparrow$ \\
\midrule
Qwen3-VL-8B & $\sim$10–15s & 50 & \textbf{19.4} & \textbf{1.32} & 57.4 & 31.2 & 29.1 & 27.6 \\
VLM-3R      & $\sim$10–15s & 50 & 24.8 & 2.10 & 60.9 & 40.3 & 42.6 & 43.1 \\
\textbf{\modelname{}$_{\mathrm{UC\text{-}SSRL}}$} & $\sim$10–15s & 50 & 22.0 & 1.65 & \textbf{76.6} & \textbf{53.0} & \textbf{57.5} & \textbf{58.1} \\
\midrule
Qwen3-VL-8B & $\sim$20–30s & 100 & 36.5 & 3.05 & 58.2 & 32.1 & 29.8 & 28.5 \\
VLM-3R      & $\sim$20–30s & 100 & 42.0 & 4.20 & 68.0 & 42.0 & 48.2 & 49.1 \\
\textbf{\modelname{}$_{\mathrm{UC\text{-}SSRL}}$} & $\sim$20–30s & 100 & \textbf{28.5} & \textbf{2.25} & \textbf{71.6} & \textbf{45.8} & \textbf{53.1} & \textbf{53.8} \\
\midrule
Qwen3-VL-8B & $\sim$120–180s & 200 & 71.2 & 7.45 & 57.8 & 31.5 & 29.3 & 27.9 \\
VLM-3R      & $\sim$120–180s & 200 & 78.5 & 10.80 & 69.5 & 43.5 & 51.1 & 52.3 \\
\textbf{\modelname{}$_{\mathrm{UC\text{-}SSRL}}$} & $\sim$120–180s & 200 & \textbf{35.0} & \textbf{3.10} & \textbf{72.4} & \textbf{46.8} & \textbf{54.4} & \textbf{55.1} \\
\bottomrule
\end{tabular}}
\vspace{-0.7cm}
\end{table}

\subsection{Efficiency and Scalability Analysis}
Beyond accuracy, we evaluate peak GPU memory, inference time, and accuracy under increasing input budgets.
All measurements are conducted on the same A100 80GB GPU, as shown in Tab.~\ref{tab:efficiency}.
At the 50-frame setting, \modelname{} achieves substantially higher accuracy than Qwen3-VL-8B and VLM-3R, while requiring less inference time than VLM-3R.
As the input budget increases, VLM-3R accumulates dense evidence and its memory and latency grow rapidly.
At 200 frames, VLM-3R requires 78.5GB memory and 10.80s inference time, whereas \modelname{} uses only 35.0GB and 3.10s while achieving higher accuracy on VSI, OSI, and MMSI.
These results show that Geometry-Consistent Memory improves end-to-end long-video scalability, not only retained-entry counts, by selectively organizing spatial evidence instead of densely retaining all observations.

\subsection{Navigation-Oriented Validation}
To connect our video spatial reasoning setting with the navigation-oriented motivation, we evaluate Habitat ObjectNav next-waypoint prediction.
Given egocentric history and a navigation query, the model predicts the next direction, testing navigation-oriented spatial perception rather than end-to-end embodied control.
Tab.~\ref{tab:habitat_nav} shows that \modelname{} outperforms Qwen3-VL-8B, with further gains from UC-SSRL.

\begin{table}[bhtp]
\vspace{-0.8cm}
\centering
\small
\begin{minipage}[t]{0.39\textwidth}
\centering
\caption{\textbf{ObjectNav next-waypoint.}
Acc., Succ., and Dist. denote direction accuracy, distance-reduction success, and remaining distance.
}
\label{tab:habitat_nav}
\vspace{-0.35cm}
\resizebox{0.8\linewidth}{!}{
\begin{tabular}{lccc}
\toprule
\textbf{Method} & \textbf{Acc.}$\uparrow$ & \textbf{Succ.}$\uparrow$ & \textbf{Dist.}$\downarrow$ \\
\midrule
Qwen3-VL-8B & 42.6 & 31.8 & 4.72 \\
\modelname{}$_{\mathrm{SFT}}$ & 47.9 & 36.5 & 4.31 \\
\modelname{}$_{\mathrm{UC\text{-}SSRL}}$ & \textbf{50.8} & \textbf{39.7} & \textbf{4.05} \\
\bottomrule
\end{tabular}
}
\end{minipage}
\hfill
\begin{minipage}[t]{0.56\textwidth}
\centering
\caption{\textbf{Ablation on GCM.}
MMSI-Video is averaged over \textit{Sufficient-Coverage} and \textit{Uniform-50}. \#Entries denotes normalized retained memory entries.
}
\label{tab:ablation_gcm}
\vspace{-0.35cm}
\resizebox{\linewidth}{!}{
\begin{tabular}{lccc|c}
\toprule
\textbf{Variant}
& \textbf{VSI}$\uparrow$
& \textbf{OSI}$\uparrow$
& \textbf{MMSI-Video}$\uparrow$
& \textbf{\#Entries}$\downarrow$ \\
\midrule
\textbf{Full (\modelname{}${\textsc{sft}}$)}
& \textbf{71.2} & \textbf{45.8} & \textbf{52.3} & \textbf{1.0$\times$} \\
w/o GateWrite
& 69.8 & 44.5 & 47.2 & 2.3$\times$ \\
w/o Fusion
& 70.3 & 45.0 & 47.9 & 1.8$\times$ \\
w/o GeoFilter
& 69.5 & 43.9 & 46.0 & 1.0$\times$ \\
\bottomrule
\end{tabular}
}
\end{minipage}
\vspace{-0.7cm}
\end{table}

\subsection{Ablation Study}

\begin{figure}[t]
\centering
\small
\begin{minipage}[t]{0.5\textwidth}
\centering
\includegraphics[width=0.8\linewidth]{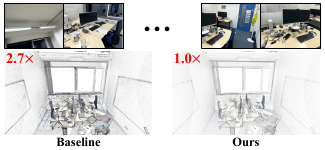}
\caption{\textbf{Qualitative effect of geometry-gated writing and fusion.}
Dense updates introduce redundant points, while GCM produces cleaner and more compact point clouds.
Red numbers denote normalized retained entries.
}
\label{fig:qual_gcm}
\end{minipage}
\hfill
\begin{minipage}[t]{0.45\textwidth}
\centering
\includegraphics[width=\linewidth]{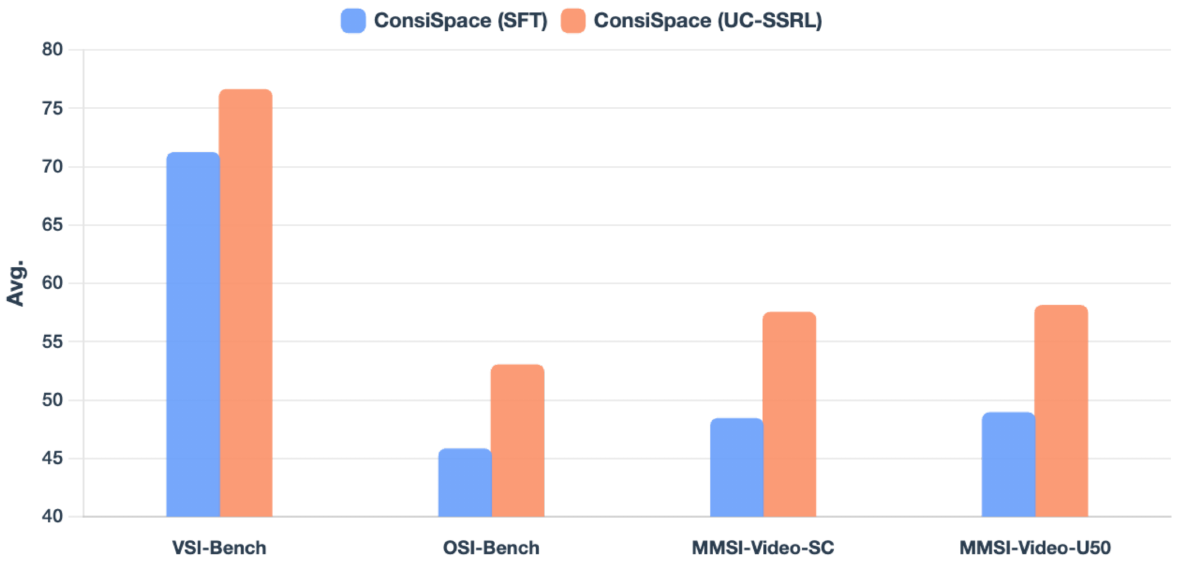}
\caption{\textbf{Effect of UC-SSRL.}
UC-SSRL improves over the SFT baseline across benchmarks, with larger gains on OSI-Bench and MMSI-Video where viewpoint changes and metric relations are more challenging.
}
\label{fig:ablation_ssrl}
\end{minipage}
\vspace{-0.7cm}
\end{figure}

\noindent\textbf{Effect of geometry-consistent memory.}
Tab.~\ref{tab:ablation_gcm} and Fig.~\ref{fig:qual_gcm} evaluate the GCM evidence lifecycle.
Removing geometry-gated writing increases retained entries to $2.3\times$, while removing geometry-consistent fusion increases them to $1.8\times$; both variants reduce accuracy, showing the importance of suppressing and merging redundant observations.
Removing geometry-filtered retrieval keeps the entry count unchanged but degrades OSI and MMSI-Video, indicating that retrieval quality depends on filtering geometrically inconsistent evidence under viewpoint changes.
Qualitatively, GCM produces cleaner and more compact point clouds than dense updates.
Overall, writing, fusion, and retrieval are complementary for compact and reliable spatial evidence organization.

\noindent\textbf{Effect of the UC-SSRL.}
Fig.~\ref{fig:ablation_ssrl} evaluates UC-SSRL starting from \modelname{}$_\textsc{sft}$.
UC-SSRL consistently improves all benchmarks, with larger gains on OSI-Bench and MMSI-Video where viewpoint changes and metric relations are more challenging.
Since UC-SSRL is initialized from the SFT model, these gains indicate that consistency rewards refine viewpoint stability rather than learning task correctness from scratch.

\begin{table}[bhtp]
\vspace{-0.8cm}
\centering
\small
\begin{minipage}[t]{0.48\textwidth}
\centering
\caption{\textbf{UC-SSRL reward ablation.}
We ablate answer-level, metric, and topological consistency rewards.
}
\label{tab:ssrl_reward_ablation}
\vspace{-0.35cm}
\resizebox{\linewidth}{!}{
\begin{tabular}{ccc|cccc}
\toprule
$R_{\mathrm{KL}}$ & $R_{\mathrm{metric}}$ & $R_{\mathrm{topo}}$
& VSI$\uparrow$ & OSI$\uparrow$ & MMSI-SC$\uparrow$ & MMSI-U50$\uparrow$ \\
\midrule
\xmark & \xmark & \xmark & 71.2 & 45.8 & 48.4 & 48.9 \\
\cmark & \xmark & \xmark & 73.5 & 48.2 & 52.1 & 52.6 \\
\xmark & \cmark & \xmark & 72.8 & 49.0 & 51.4 & 51.9 \\
\xmark & \xmark & \cmark & 72.9 & 47.1 & 51.7 & 52.2 \\
\cmark & \cmark & \xmark & 75.1 & 51.6 & 55.4 & 56.0 \\
\cmark & \xmark & \cmark & 75.4 & 50.3 & 55.7 & 56.4 \\
\xmark & \cmark & \cmark & 74.5 & 50.8 & 54.8 & 55.5 \\
\midrule
\cmark & \cmark & \cmark & \textbf{76.6} & \textbf{53.0} & \textbf{57.5} & \textbf{58.1} \\
\bottomrule
\end{tabular}}
\end{minipage}
\hfill
\begin{minipage}[t]{0.48\textwidth}
\centering
\caption{\textbf{Robustness to noisy VGGT geometry.}
We perturb VGGT geometry before GCM operations.
}
\label{tab:geo_noise}
\vspace{-0.35cm}
\resizebox{0.83\linewidth}{!}{
\begin{tabular}{lccccc}
\toprule
\textbf{Method} & \textbf{Noise} & VSI$\uparrow$ & OSI$\uparrow$ & SC$\uparrow$ & U50$\uparrow$ \\
\midrule
GeoThinker & Clean & 72.6 & 44.5 & 48.2 & 49.1 \\
\modelname{}${_\mathrm{UC\text{-}SSRL}}$ & Clean & \textbf{76.6} & \textbf{53.0} & \textbf{57.5} & \textbf{58.1} \\ \midrule
GeoThinker & 10\% & 70.8 & 42.1 & 45.3 & 46.5 \\
\modelname{}${_\mathrm{UC\text{-}SSRL}}$ & 10\% & \textbf{75.9} & \textbf{52.4} & \textbf{56.8} & \textbf{57.4} \\ \midrule
GeoThinker & 20\% & 68.4 & 38.8 & 41.6 & 42.8 \\
\modelname{}${_\mathrm{UC\text{-}SSRL}}$ & 20\% & \textbf{74.9} & \textbf{51.3} & \textbf{55.6} & \textbf{56.1} \\ \midrule
GeoThinker & 30\% & 65.2 & 35.1 & 38.2 & 39.4 \\
\modelname{}${_\mathrm{UC\text{-}SSRL}}$ & 30\% & \textbf{73.1} & \textbf{49.6} & \textbf{53.4} & \textbf{54.3} \\
\bottomrule
\end{tabular}}
\end{minipage}
\vspace{-0.75cm}
\end{table}

\noindent\textbf{UC-SSRL reward ablation.}
Tab.~\ref{tab:ssrl_reward_ablation} ablates the three UC-SSRL rewards.
Each reward improves over SFT, pairwise combinations bring further gains, and the full objective performs best across VSI, OSI, and MMSI-Video.
This shows that answer-level, metric, and topological consistency are complementary, and that UC-SSRL refines cross-view stability from an SFT-initialized model rather than learning correctness from consistency alone.

\noindent\textbf{Robustness to noisy geometry.}
Tab.~\ref{tab:geo_noise} evaluates robustness to perturbed VGGT geometry.
We add noise to depth, translation, and rotation cues before GCM operations while keeping RGB inputs and questions unchanged.
Performance degrades gradually as noise increases, but \modelname{} remains consistently stronger than GeoThinker, suggesting that GCM benefits from geometry cues without requiring perfect geometry estimates.

\begin{wraptable}{r}{0.5\textwidth}
\vspace{-0.9cm}
\centering
\small
\caption{\textbf{Readout strategy ablation.}
We compare summary-only, Top-$K$-only, and their combination under the same backbone and memory budget.
}
\resizebox{0.5\textwidth}{!}{
\begin{tabular}{lcccc}
\toprule
\textbf{Variant} 
& \textbf{VSI} $\uparrow$ 
& \textbf{OSI} $\uparrow$ 
& \textbf{MMSI-SC} $\uparrow$ 
& \textbf{MMSI-U50} $\uparrow$ \\
\midrule
\textbf{Summary + Top-$K$}  & \textbf{71.2} & \textbf{45.8} & \textbf{51.8} & \textbf{52.9} \\
Summary-only           & 69.8 & 43.9 & 48.7 & 47.1 \\
Top-$K$-only              & 70.9 & 46.2 & 49.3 & 50.9 \\
\bottomrule
\end{tabular}
}
\vspace{-0.8cm}
\label{tab:readout_ablation}
\end{wraptable}
\noindent\textbf{Summary vs.\ Top-$K$ readout.}
Tab.~\ref{tab:readout_ablation} ablates the readout strategy from the Geometry-Consistent Memory.
Summary-only readout lacks fine-grained local evidence, while Top-$K$-only readout misses global spatial context.
Combining summary tokens with Top-$K$ retrieval gives the best overall performance, showing that global summarization and sparse evidence selection are complementary.

\begin{table}[tbhp]
\vspace{-0.6cm}
\centering
\small
\begin{minipage}[t]{0.4\textwidth}
\centering
\caption{\textbf{Sensitivity to writing thresholds.}
VSI and OSI averages of \modelname{}$_{\textsc{sft}}$ under different $(\tau_p,\tau_\theta)$.
The default setting is $(0.15\,\mathrm{m},15^\circ)$.
}
\label{tab:write_threshold}
\vspace{-0.35cm}
\resizebox{0.7\linewidth}{!}{
\begin{tabular}{cc|cc}
\toprule
$\tau_p$ (m) & $\tau_\theta$ (deg) & VSI$\uparrow$ & OSI$\uparrow$ \\
\midrule
0.10 & 15 & 71.0 & 45.6 \\
0.15 & 15 & \textbf{71.2} & \textbf{45.8} \\
0.25 & 15 & 70.9 & 45.3 \\
\midrule
0.15 & 10 & 71.1 & 45.7 \\
0.15 & 15 & \textbf{71.2} & \textbf{45.8} \\
0.15 & 30 & 70.8 & 45.4 \\
\bottomrule
\end{tabular}
}
\end{minipage}
\hfill
\begin{minipage}[t]{0.55\textwidth}
\centering
\caption{\textbf{Sensitivity to fusion thresholds.}
VSI and OSI averages under different fusion neighborhoods $(r,\theta)$.
\#Entries denotes retained entries normalized by the default $(0.40\,\mathrm{m},30^\circ)$.
}
\label{tab:fusion_threshold}
\vspace{-0.35cm}
\resizebox{0.78\linewidth}{!}{
\begin{tabular}{cc|ccc}
\toprule
$r$ (m) & $\theta$ (deg) & VSI$\uparrow$ & OSI$\uparrow$ & \#Entries$\downarrow$ \\
\midrule
0.20 & 30 & 71.0 & 45.5 & 1.18$\times$ \\
0.40 & 30 & \textbf{71.2} & \textbf{45.8} & 1.00$\times$ \\
0.60 & 30 & 70.9 & 45.4 & 0.92$\times$ \\
\midrule
0.40 & 15 & 71.1 & 45.6 & 1.10$\times$ \\
0.40 & 30 & \textbf{71.2} & \textbf{45.8} & 1.00$\times$ \\
0.40 & 45 & 70.8 & 45.3 & 0.94$\times$ \\
\bottomrule
\end{tabular}
}
\end{minipage}
\vspace{-0.8cm}
\end{table}

\noindent\textbf{Threshold sensitivity.}
Tab.~\ref{tab:write_threshold} and Tab.~\ref{tab:fusion_threshold} evaluate the sensitivity of writing thresholds $(\tau_p,\tau_\theta)$ and fusion neighborhood thresholds $(r,\theta)$.
Around the default settings, VSI and OSI vary only slightly, indicating that both geometry-gated writing and geometry-consistent fusion are robust to threshold choices.
For fusion, the retained entries change as expected: smaller neighborhoods retain more entries, while larger neighborhoods enable more aggressive fusion.
Overall, \modelname{} exhibits a broad stable regime and does not rely on finely tuned geometric thresholds.

\subsection{Qualitative Results}

\begin{figure}[t]
    \centering
    \small
    \includegraphics[width=0.9\linewidth]{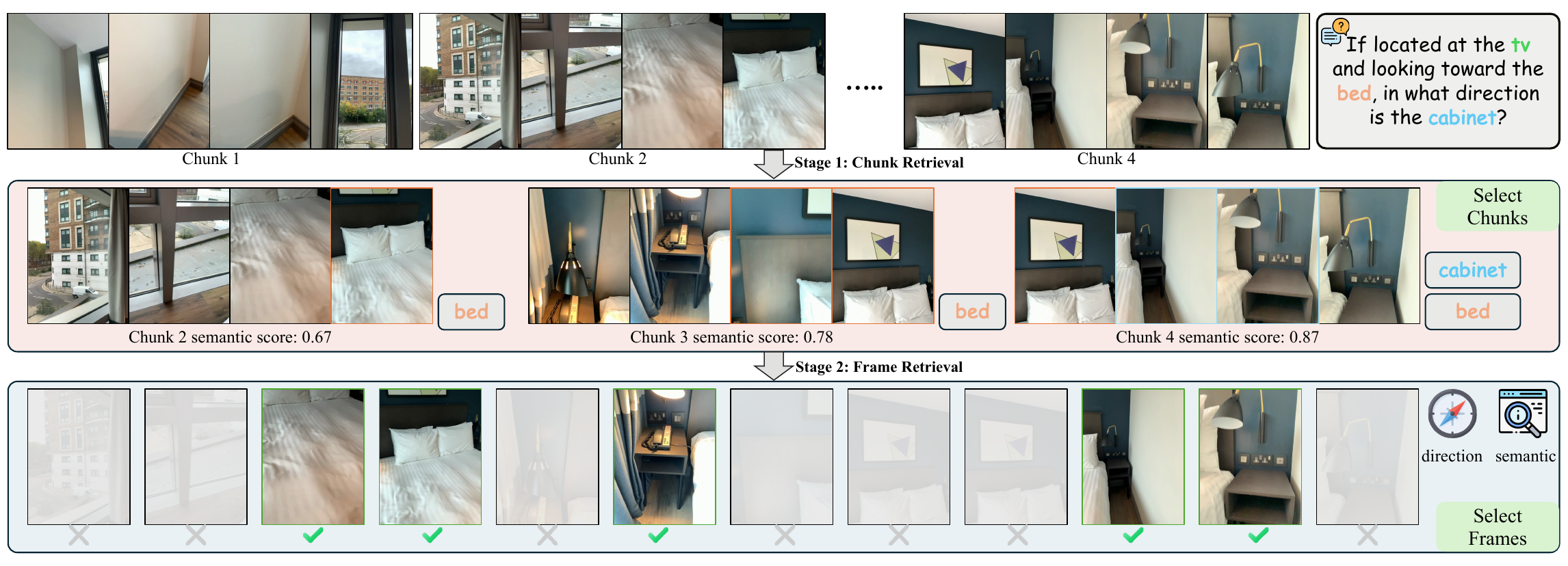}
    \caption{\textbf{Qualitative visualization of geometry-filtered retrieval.}
    GCM first retrieves semantically relevant chunks and then selects frames using direction consistency.
    Gray~\ding{55} denotes suppressed frames, and green~\ding{51} denotes selected evidence.
    }
    \label{fig:qual_retrieval}
    \vspace{-0.4cm}
\end{figure}

\noindent\textbf{Geometry-filtered retrieval.}
Fig.~\ref{fig:qual_retrieval} visualizes the two-stage retrieval process.
After chunk-level semantic retrieval, GCM performs frame-level selection with a direction-consistency cue.
Frames that match the query semantics but conflict with the inferred viewpoint are suppressed.

\begin{figure}[t]
    \centering
    \includegraphics[width=0.9\linewidth]{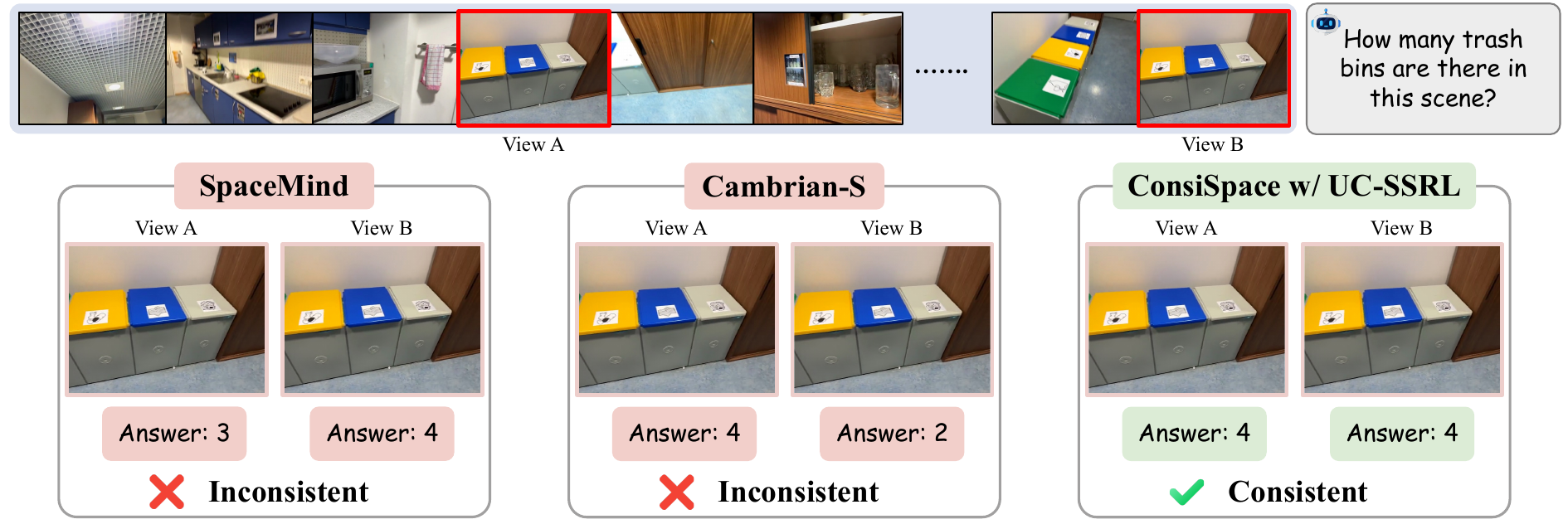}
    \caption{\textbf{Consistency under revisit views.}
    Given the same query, baseline models produce inconsistent counting answers across two revisited viewpoints, while \modelname{} with UC-SSRL gives consistent predictions.
    }
    \label{fig:qual_loop}
    \vspace{-0.75cm}
\end{figure}

\noindent\textbf{Consistency under revisit views.}
Fig.~\ref{fig:qual_loop} shows a revisit-view example for UC-SSRL.
Given the same video and query, baseline models produce different answers across nearby viewpoints, indicating unstable spatial grounding.
In contrast, \modelname{} with UC-SSRL gives consistent predictions across View A and View B.
This suggests that UC-SSRL refines viewpoint stability after SFT, rather than learning task correctness from consistency alone.

\noindent\textbf{Additional visualizations.}
We provide more qualitative examples on VSI-Bench, MMSI-Video-Bench, and OSI-Bench in the \textit{Supp. Mat.}~\ref{sec:visualization}.

\section{Conclusion}
In this paper, we propose \textbf{\modelname{}}, a geometry-consistent framework for video spatial reasoning.
\modelname{} builds a Geometry-Consistent Memory on top of generic video encoders and VLMs, and manages evidence via geometry-gated writing, geometry-consistent fusion, and geometry-filtered retrieval to reduce redundancy and retrieve query-relevant spatial cues.
We further introduce a Unified Consistency SSRL objective that encourages cross-view spatial agreement after supervised fine-tuning, improving viewpoint stability without additional human annotations.
Experiments on three challenging video spatial reasoning benchmarks show that \modelname{} consistently outperforms strong prior methods.
We hope our results highlight geometric consistency as a simple and effective principle for robust video spatial reasoning.

\noindent\textbf{Acknowledgements.}
This work was supported by the NSFC under Grant No. 62376121, Basic Research Program of Jiangsu under Grant No. BK20251999, Gusu Innovation Leading Talent Program under Grant No. ZXL2025319, and Jiangsu Provincial Science \& Technology Major Project under Grant No. BG2024042.

%
%
\bibliographystyle{splncs04}
\bibliography{main}

@article{ouyang2025spacer,
  title={SpaceR: Reinforcing MLLMs in Video Spatial Reasoning},
  author={Ouyang, Kun and Liu, Yuanxin and Wu, Haoning and Liu, Yi and Zhou, Hao and Zhou, Jie and Meng, Fandong and Sun, Xu},
  journal={arXiv preprint arXiv:2504.01805},
  year={2025}
}

@article{wu2025reinforcing,
  title={Reinforcing spatial reasoning in vision-language models with interwoven thinking and visual drawing},
  author={Wu, Junfei and Guan, Jian and Feng, Kaituo and Liu, Qiang and Wu, Shu and Wang, Liang and Wu, Wei and Tan, Tieniu},
  journal={arXiv preprint arXiv:2506.09965},
  year={2025}
}

@article{wu2025spatial,
  title={Spatial-mllm: Boosting mllm capabilities in visual-based spatial intelligence},
  author={Wu, Diankun and Liu, Fangfu and Hung, Yi-Hsin and Duan, Yueqi},
  journal={arXiv preprint arXiv:2505.23747},
  year={2025}
}

@article{fan2025vlm,
  title={VLM-3R: Vision-Language Models Augmented with Instruction-Aligned 3D Reconstruction},
  author={Fan, Zhiwen and Zhang, Jian and Li, Renjie and Zhang, Junge and Chen, Runjin and Hu, Hezhen and Wang, Kevin and Qu, Huaizhi and Wang, Dilin and Yan, Zhicheng and others},
  journal={arXiv preprint arXiv:2505.20279},
  year={2025}
}

@article{zhao2025spacemind,
  title={SpaceMind: Camera-Guided Modality Fusion for Spatial Reasoning in Vision-Language Models},
  author={Zhao, Ruosen and Zhang, Zhikang and Xu, Jialei and Chang, Jiahao and Chen, Dong and Li, Lingyun and Sun, Weijian and Wei, Zizhuang},
  journal={arXiv preprint arXiv:2511.23075},
  year={2025}
}

@article{gao2026map2thought,
  title={Map2Thought: Explicit 3D Spatial Reasoning via Metric Cognitive Maps},
  author={Gao, Xiangjun and Zhang, Zhensong and Chen, Dave Zhenyu and Xu, Songcen and Quan, Long and P{\'e}rez-Pellitero, Eduardo and Jang, Youngkyoon},
  journal={arXiv preprint arXiv:2601.11442},
  year={2026}
}

@article{bai2025qwen3,
  title={Qwen3-vl technical report},
  author={Bai, Shuai and Cai, Yuxuan and Chen, Ruizhe and Chen, Keqin and Chen, Xionghui and Cheng, Zesen and Deng, Lianghao and Ding, Wei and Gao, Chang and Ge, Chunjiang and others},
  journal={arXiv preprint arXiv:2511.21631},
  year={2025}
}

@article{team2023gemini,
  title={Gemini: a family of highly capable multimodal models},
  author={Team, Gemini and Anil, Rohan and Borgeaud, Sebastian and Alayrac, Jean-Baptiste and Yu, Jiahui and Soricut, Radu and Schalkwyk, Johan and Dai, Andrew M and Hauth, Anja and Millican, Katie and others},
  journal={arXiv preprint arXiv:2312.11805},
  year={2023}
}

@article{singh2025openai,
  title={Openai gpt-5 system card},
  author={Singh, Aaditya and Fry, Adam and Perelman, Adam and Tart, Adam and Ganesh, Adi and El-Kishky, Ahmed and McLaughlin, Aidan and Low, Aiden and Ostrow, AJ and Ananthram, Akhila and others},
  journal={arXiv preprint arXiv:2601.03267},
  year={2025}
}

@inproceedings{dongfang2026multimodal,
  title={Are multimodal large language models ready for omnidirectional spatial reasoning?},
  author={Dongfang, Zihao and Zheng, Xu and Weng, Ziqiao and Lyu, Yuanhuiyi and Paudel, Danda Pani and Van Gool, Luc and Yang, Kailun and Hu, Xuming},
  booktitle={Proceedings of the IEEE/CVF Conference on Computer Vision and Pattern Recognition},
  pages={9759--9769},
  year={2026}
}

@article{li2025viewspatial,
  title={Viewspatial-bench: Evaluating multi-perspective spatial localization in vision-language models},
  author={Li, Dingming and Li, Hongxing and Wang, Zixuan and Yan, Yuchen and Zhang, Hang and Chen, Siqi and Hou, Guiyang and Jiang, Shengpei and Zhang, Wenqi and Shen, Yongliang and others},
  journal={arXiv preprint arXiv:2505.21500},
  year={2025}
}

@article{zhang2026flatland,
  title={From flatland to space: Teaching vision-language models to perceive and reason in 3d},
  author={Zhang, Jiahui and Chen, Yurui and Xu, Yueming and Huang, Ze and Mei, Jilin and Chen, Chunhui and Zhou, Yanpeng and Yuan, Yu-Jie and Cai, Xinyue and Huang, Guowei and others},
  journal={Advances in Neural Information Processing Systems},
  volume={38},
  year={2026}
}

@inproceedings{yang2025thinking,
  title={Thinking in space: How multimodal large language models see, remember, and recall spaces},
  author={Yang, Jihan and Yang, Shusheng and Gupta, Anjali W and Han, Rilyn and Fei-Fei, Li and Xie, Saining},
  booktitle={Proceedings of the Computer Vision and Pattern Recognition Conference},
  pages={10632--10643},
  year={2025}
}

@article{lin2025mmsi,
  title={MMSI-Video-Bench: A Holistic Benchmark for Video-Based Spatial Intelligence},
  author={Lin, Jingli and Xu, Runsen and Zhu, Shaohao and Yang, Sihan and Cao, Peizhou and Ran, Yunlong and Hu, Miao and Zhu, Chenming and Xie, Yiman and Long, Yilin and others},
  journal={arXiv preprint arXiv:2512.10863},
  year={2025}
}

@inproceedings{zheng2025video,
  title={Video-3d llm: Learning position-aware video representation for 3d scene understanding},
  author={Zheng, Duo and Huang, Shijia and Wang, Liwei},
  booktitle={Proceedings of the IEEE/CVF Conference on Computer Vision and Pattern Recognition},
  pages={8995--9006},
  year={2025}
}

@inproceedings{wang2025ross3d,
  title={Ross3d: Reconstructive visual instruction tuning with 3d-awareness},
  author={Wang, Haochen and Zhao, Yucheng and Wang, Tiancai and Fan, Haoqiang and Zhang, Xiangyu and Zhang, Zhaoxiang},
  booktitle={Proceedings of the IEEE/CVF International Conference on Computer Vision},
  pages={9275--9286},
  year={2025}
}

@article{wang2025n3d,
  title={N3D-VLM: Native 3D Grounding Enables Accurate Spatial Reasoning in Vision-Language Models},
  author={Wang, Yuxin and Ke, Lei and Zhang, Boqiang and Qu, Tianyuan and Yu, Hanxun and Huang, Zhenpeng and Yu, Meng and Xu, Dan and Yu, Dong},
  journal={arXiv preprint arXiv:2512.16561},
  year={2025}
}

@article{yang2025cambrian,
  title={Cambrian-s: Towards spatial supersensing in video},
  author={Yang, Shusheng and Yang, Jihan and Huang, Pinzhi and Brown, Ellis and Yang, Zihao and Yu, Yue and Tong, Shengbang and Zheng, Zihan and Xu, Yifan and Wang, Muhan and others},
  journal={arXiv preprint arXiv:2511.04670},
  year={2025}
}

@article{li2025spatialladder,
  title={Spatialladder: Progressive training for spatial reasoning in vision-language models},
  author={Li, Hongxing and Li, Dingming and Wang, Zixuan and Yan, Yuchen and Wu, Hang and Zhang, Wenqi and Shen, Yongliang and Lu, Weiming and Xiao, Jun and Zhuang, Yueting},
  journal={arXiv preprint arXiv:2510.08531},
  year={2025}
}

@inproceedings{
chen2026reasoning,
title={Reasoning in Space via Grounding in the World},
author={Yiming Chen and Zekun Qi and Wenyao Zhang and Xin Jin and Li Zhang and Peidong Liu},
booktitle={The Fourteenth International Conference on Learning Representations},
year={2026}
}

@inproceedings{wang2025vggt,
  title={Vggt: Visual geometry grounded transformer},
  author={Wang, Jianyuan and Chen, Minghao and Karaev, Nikita and Vedaldi, Andrea and Rupprecht, Christian and Novotny, David},
  booktitle={Proceedings of the Computer Vision and Pattern Recognition Conference},
  pages={5294--5306},
  year={2025}
}

@inproceedings{
wang2026pi,
title={$\pi^3$: Permutation-Equivariant Visual Geometry Learning},
author={Yifan Wang and Jianjun Zhou and Haoyi Zhu and Wenzheng Chang and Yang Zhou and Zizun Li and Junyi Chen and Jiangmiao Pang and Chunhua Shen and Tong He},
booktitle={The Fourteenth International Conference on Learning Representations},
year={2026}
}

@article{lin2025depth,
  title={Depth anything 3: Recovering the visual space from any views},
  author={Lin, Haotong and Chen, Sili and Liew, Junhao and Chen, Donny Y and Li, Zhenyu and Shi, Guang and Feng, Jiashi and Kang, Bingyi},
  journal={arXiv preprint arXiv:2511.10647},
  year={2025}
}

@article{zheng2025learning,
  title={Learning from videos for 3d world: Enhancing mllms with 3d vision geometry priors},
  author={Zheng, Duo and Huang, Shijia and Li, Yanyang and Wang, Liwei},
  journal={arXiv preprint arXiv:2505.24625},
  year={2025}
}

@inproceedings{hu2025g,
  title={G$^2$VLM: Geometry Grounded Vision Language Model with Unified 3D Reconstruction and Spatial Reasoning},
  author={Hu, Wenbo and Lin, Jingli and Long, Yilin and Ran, Yunlong and Jiang, Lihan and Wang, Yifan and Zhu, Chenming and Xu, Runsen and Wang, Tai and Pang, Jiangmiao},
  booktitle={Proceedings of the IEEE/CVF Conference on Computer Vision and Pattern Recognition},
  pages={9535--9546},
  year={2026}
}

@article{li2026thinking,
  title={Thinking with Geometry: Active Geometry Integration for Spatial Reasoning},
  author={Li, Haoyuan and Cao, Qihang and Tang, Tao and Xiang, Kun and Guo, Zihan and Han, Jianhua and Xu, Hang and Liang, Xiaodan},
  journal={arXiv preprint arXiv:2602.06037},
  year={2026}
}

@article{zhang2026think3d,
  title={Think3D: Thinking with Space for Spatial Reasoning},
  author={Zhang, Zaibin and Wu, Yuhan and Jia, Lianjie and Wang, Yifan and Zhang, Zhongbo and Li, Yijiang and Ran, Binghao and Zhang, Fuxi and Sun, Zhuohan and Yin, Zhenfei and others},
  journal={arXiv preprint arXiv:2601.13029},
  year={2026}
}

@article{huang20253drs,
  title={3drs: Mllms need 3d-aware representation supervision for scene understanding},
  author={Huang, Xiaohu and Wu, Jingjing and Xie, Qunyi and Han, Kai},
  journal={arXiv preprint arXiv:2506.01946},
  year={2025}
}

@article{li2025spatial,
  title={Spatial forcing: Implicit spatial representation alignment for vision-language-action model},
  author={Li, Fuhao and Song, Wenxuan and Zhao, Han and Wang, Jingbo and Ding, Pengxiang and Wang, Donglin and Zeng, Long and Li, Haoang},
  journal={arXiv preprint arXiv:2510.12276},
  year={2025}
}

@inproceedings{krantz2020beyond,
  title={Beyond the nav-graph: Vision-and-language navigation in continuous environments},
  author={Krantz, Jacob and Wijmans, Erik and Majumdar, Arjun and Batra, Dhruv and Lee, Stefan},
  booktitle={European Conference on Computer Vision},
  pages={104--120},
  year={2020},
  organization={Springer}
}

@inproceedings{savva2019habitat,
  title={Habitat: A platform for embodied ai research},
  author={Savva, Manolis and Kadian, Abhishek and Maksymets, Oleksandr and Zhao, Yili and Wijmans, Erik and Jain, Bhavana and Straub, Julian and Liu, Jia and Koltun, Vladlen and Malik, Jitendra and others},
  booktitle={Proceedings of the IEEE/CVF international conference on computer vision},
  pages={9339--9347},
  year={2019}
}

@article{liu2025nav,
  title={Nav-r1: Reasoning and navigation in embodied scenes},
  author={Liu, Qingxiang and Huang, Ting and Zhang, Zeyu and Tang, Hao},
  journal={arXiv preprint arXiv:2509.10884},
  year={2025}
}

@article{huang2025mobilevla,
  title={Mobilevla-r1: Reinforcing vision-language-action for mobile robots},
  author={Huang, Ting and Li, Dongjian and Yang, Rui and Zhang, Zeyu and Yang, Zida and Tang, Hao},
  journal={arXiv preprint arXiv:2511.17889},
  year={2025}
}

@inproceedings{wang2025lvbench,
  title={Lvbench: An extreme long video understanding benchmark},
  author={Wang, Weihan and He, Zehai and Hong, Wenyi and Cheng, Yean and Zhang, Xiaohan and Qi, Ji and Ding, Ming and Gu, Xiaotao and Huang, Shiyu and Xu, Bin and others},
  booktitle={Proceedings of the IEEE/CVF International Conference on Computer Vision},
  pages={22958--22967},
  year={2025}
}

@inproceedings{
chen2025cgbench,
title={{CG}-Bench: Clue-grounded Question Answering Benchmark for Long Video Understanding},
author={Guo Chen and Yicheng Liu and Yifei Huang and Baoqi Pei and Jilan Xu and Yuping He and Tong Lu and Yali Wang and Limin Wang},
booktitle={The Thirteenth International Conference on Learning Representations},
year={2025}
}

@article{huang20253d,
  title={3d-r1: Enhancing reasoning in 3d vlms for unified scene understanding},
  author={Huang, Ting and Zhang, Zeyu and Tang, Hao},
  journal={arXiv preprint arXiv:2507.23478},
  year={2025}
}

@inproceedings{
huang20253dcoca,
title={3D CoCa: Contrastive Learners are 3D Captioners},
author={Ting Huang and Zeyu Zhang and Yemin Wang and Hao Tang},
booktitle={Thirteenth International Conference on 3D Vision},
year={2026}
}

@article{huang2025dc,
  title={DC-Scene: Data-Centric Learning for 3D Scene Understanding},
  author={Huang, Ting and Zhang, Zeyu and Zhang, Ruicheng and Zhao, Yang},
  journal={arXiv preprint arXiv:2505.15232},
  year={2025}
}

@article{tang20263d,
  title={3D CoCa v2: Contrastive Learners with Test-Time Search for Generalizable Spatial Intelligence},
  author={Tang, Hao and Huang, Ting and Zhang, Zeyu},
  journal={arXiv preprint arXiv:2601.06496},
  year={2026}
}

@article{wu2025indoor,
  title={From indoor to open world: Revealing the spatial reasoning gap in mllms},
  author={Wu, Mingrui and Wang, Zhaozhi and Wang, Fangjinhua and Yang, Jiaolong and Pollefeys, Marc and Zhang, Tong},
  journal={arXiv preprint arXiv:2512.19683},
  year={2025}
}

@inproceedings{zhi2025lscenellm,
  title={Lscenellm: Enhancing large 3d scene understanding using adaptive visual preferences},
  author={Zhi, Hongyan and Chen, Peihao and Li, Junyan and Ma, Shuailei and Sun, Xinyu and Xiang, Tianhang and Lei, Yinjie and Tan, Mingkui and Gan, Chuang},
  booktitle={Proceedings of the Computer Vision and Pattern Recognition Conference},
  pages={3761--3771},
  year={2025}
}

@article{tschannen2025siglip,
  title={Siglip 2: Multilingual vision-language encoders with improved semantic understanding, localization, and dense features},
  author={Tschannen, Michael and Gritsenko, Alexey and Wang, Xiao and Naeem, Muhammad Ferjad and Alabdulmohsin, Ibrahim and Parthasarathy, Nikhil and Evans, Talfan and Beyer, Lucas and Xia, Ye and Mustafa, Basil and others},
  journal={arXiv preprint arXiv:2502.14786},
  year={2025}
}

@inproceedings{ji2025robobrain,
  title={Robobrain: A unified brain model for robotic manipulation from abstract to concrete},
  author={Ji, Yuheng and Tan, Huajie and Shi, Jiayu and Hao, Xiaoshuai and Zhang, Yuan and Zhang, Hengyuan and Wang, Pengwei and Zhao, Mengdi and Mu, Yao and An, Pengju and others},
  booktitle={Proceedings of the IEEE/CVF Conference on Computer Vision and Pattern Recognition},
  pages={1724--1734},
  year={2025}
}

@inproceedings{wang2024dust3r,
  title={Dust3r: Geometric 3d vision made easy},
  author={Wang, Shuzhe and Leroy, Vincent and Cabon, Yohann and Chidlovskii, Boris and Revaud, Jerome},
  booktitle={Proceedings of the IEEE/CVF conference on computer vision and pattern recognition},
  pages={20697--20709},
  year={2024}
}

@inproceedings{leroy2024grounding,
  title={Grounding image matching in 3d with mast3r},
  author={Leroy, Vincent and Cabon, Yohann and Revaud, J{\'e}r{\^o}me},
  booktitle={European conference on computer vision},
  pages={71--91},
  year={2024},
  organization={Springer}
}

@article{liu2025spatial,
  title={Spatial-ssrl: Enhancing spatial understanding via self-supervised reinforcement learning},
  author={Liu, Yuhong and Zhang, Beichen and Zang, Yuhang and Cao, Yuhang and Xing, Long and Dong, Xiaoyi and Duan, Haodong and Lin, Dahua and Wang, Jiaqi},
  journal={arXiv preprint arXiv:2510.27606},
  year={2025}
}

@article{nuscenes2019,
  title={nuScenes: A multimodal dataset for autonomous driving},
  author={Holger Caesar and Varun Bankiti and Alex H. Lang and Sourabh Vora and 
          Venice Erin Liong and Qiang Xu and Anush Krishnan and Yu Pan and 
          Giancarlo Baldan and Oscar Beijbom},
  journal={arXiv preprint arXiv:1903.11027},
  year={2019}
}

@inproceedings{chen2025longvila,
  title={Longvila: Scaling long-context visual language models for long videos},
  author={Chen, Yukang and Xue, Fuzhao and Li, Dacheng and Hu, Qinghao and Zhu, Ligeng and Li, Xiuyu and Fang, Yunhao and Tang, Haotian and Yang, Shang and Liu, Zhijian and others},
  booktitle={International Conference on Learning Representations},
  volume={2025},
  pages={18227--18246},
  year={2025}
}

@inproceedings{wang2024videoagent,
  title={Videoagent: Long-form video understanding with large language model as agent},
  author={Wang, Xiaohan and Zhang, Yuhui and Zohar, Orr and Yeung-Levy, Serena},
  booktitle={European Conference on Computer Vision},
  pages={58--76},
  year={2024},
  organization={Springer}
}

@article{ramakrishnan2021habitat,
  title={Habitat-matterport 3d dataset (hm3d): 1000 large-scale 3d environments for embodied ai},
  author={Ramakrishnan, Santhosh K and Gokaslan, Aaron and Wijmans, Erik and Maksymets, Oleksandr and Clegg, Alex and Turner, John and Undersander, Eric and Galuba, Wojciech and Westbury, Andrew and Chang, Angel X and others},
  journal={arXiv preprint arXiv:2109.08238},
  year={2021}
}

@article{wang2026let,
  title={Let Geometry GUIDE: Layer-wise Unrolling of Geometric Priors in Multimodal LLMs},
  author={Wang, Chongyu and Huang, Ting and Sun, Chunyu and Ning, Xinyu and Wang, Di and Tang, Hao},
  journal={arXiv preprint arXiv:2604.05695},
  year={2026}
}

@article{zhang2026multigranularity,
  title={Multigranularity-3DQA: Dynamic Multi-Granularity Perception and Gated Dual-Stage Encoder-Decoder for 3D Question Answering},
  author={Zhang, Letao and Wan, Weibing and Huang, Ting and Fang, Zhijun and Zhang, Hongbing},
  journal={Expert Systems with Applications},
  pages={131676},
  year={2026},
  publisher={Elsevier}
}

@article{wang2026lamp,
  title={LaMP: Learning Vision-Language-Action Policies with 3D Scene Flow as Latent Motion Prior},
  author={Wang, Xinkai and Wang, Chenyi and Xu, Yifu and Ye, Mingzhe and Zhang, Fu-Cheng and Tian, Jialin and Zhan, Xinyu and Zhu, Lifeng and Lu, Cewu and Yang, Lixin},
  journal={arXiv preprint arXiv:2603.25399},
  year={2026}
}

\clearpage

\appendix

\section{Outdoor Data Synthesis}
\label{sec:supp_data_synthesis}
This section describes our automated pipeline for synthesizing \textbf{nuScenes-10K}, an outdoor spatial-reasoning instruction set from nuScenes~\cite{nuscenes2019}.
nuScenes-10K complements the indoor VSI-590K training data with outdoor geometry patterns, such as long-range depth, large-scale scenes, and dynamic agents, while keeping the supervision compatible with our video spatial reasoning formulation.

\subsection{Data Sources and Representations}
We build on nuScenes sequences with calibrated multi-camera imagery, ego poses, 3D object annotations, and optional LiDAR-based geometric verification.
For each driving clip, we sample short temporal windows and form a multi-view video input $V={I_t}_{t=1}^T$ using the same frame-budget setting as training.
We reuse the same geometry extraction stack as in the main pipeline, including pose and depth cues, so that outdoor samples follow the same input interface as \modelname{}.

\subsection{Automated Instruction and Answer Generation}
For each clip, we generate question-answer pairs covering three families:
\begin{itemize}
\item \textbf{Topological and relational:} relative direction, relative ordering, and ego-heading-conditioned spatial relations.
\item \textbf{Metric:} approximate distance, size comparison, and nearest or farthest object queries, with numeric, binned, or multiple-choice answers.
\item \textbf{Motion and interaction:} whether an agent approaches, recedes, crosses, or changes its relative state within the temporal window.
\end{itemize}
Questions are generated by templates with controlled lexical diversity.
Answers are computed from calibrated geometry: we transform 3D object centers into the ego frame, derive relative directions by azimuth with respect to the ego forward axis, and compute metric quantities by Euclidean distance with optional depth verification.
For multiple-choice questions, we construct fixed option sets and keep only samples with unambiguous labels.

\subsection{Quality Control and Verification}
We generate 22,518 candidate QA pairs and apply multi-stage filtering to improve label reliability.
Specifically, we filter candidates by geometry validity, answer parseability, consistency checking, MLLM verification, and de-duplication, yielding 10,000 final QA pairs.
Tab.~\ref{tab:nuscenes_filter} summarizes the filtering statistics.

\begin{table}[t]
\centering
\small
\begin{minipage}[t]{0.48\linewidth}
\centering
\caption{\textbf{nuScenes-10K filtering statistics.}
Multi-stage filtering reduces 22,518 generated candidates to 10,000 final QA pairs.
}
\vspace{-0.3cm}
\label{tab:nuscenes_filter}
\resizebox{0.48\linewidth}{!}{
\begin{tabular}{lcc}
\toprule
\textbf{Stage} & \textbf{\#QA} & \textbf{Ret.} \\
\midrule
Raw candidates      & 22,518 & 100\% \\
Geometry-valid      & 19,806 & 88.0\% \\
Parseable           & 17,932 & 79.6\% \\
Consistency checked & 14,516 & 64.5\% \\
MLLM pass           & 11,842 & 52.6\% \\
De-duplicated       & 10,436 & 46.3\% \\
Final               & 10,000 & 44.4\% \\
\bottomrule
\end{tabular}}
\end{minipage}
\hfill
\begin{minipage}[t]{0.48\linewidth}
\centering
\caption{\textbf{Human verification of nuScenes-10K.}
Manual verification on 1,000 random QA pairs shows 91.6\% correctness.
}
\vspace{-0.3cm}
\label{tab:nuscenes_human}
\resizebox{\linewidth}{!}{
\begin{tabular}{lrc}
\toprule
\textbf{Category} & \textbf{Ratio} & \textbf{Main reason} \\
\midrule
Correct & 91.6\% & Valid object/relation/answer \\
Ambiguous reference & 3.4\% & Multiple plausible objects \\
Occluded / distant & 2.1\% & Hard to verify visually \\
Annotation sparsity & 1.7\% & Missing/incomplete 3D box \\
Incorrect & 1.2\% & Wrong relation/metric answer \\
\bottomrule
\end{tabular}}
\end{minipage}
\vspace{-0.4cm}
\end{table}

We further check potential overlap between nuScenes-10K and OSI-Bench.
nuScenes-10K uses autonomous-driving ego-pose and 3D-box annotations, while OSI-Bench uses pedestrian-view videos with independent 3D ground truth and QA protocols.
We find no shared videos or annotations between the two datasets, with only generic object-category overlap.

To estimate label quality, we manually verify 1,000 randomly sampled QA pairs.
As shown in Tab.~\ref{tab:nuscenes_human}, 91.6\% of the verified samples are correct.
The remaining errors mainly come from ambiguous references, occlusion or distance ambiguity, sparse annotations, and incorrect metric or relation answers.
We keep the final data format identical to indoor instruction tuning, enabling mixed SFT training with VSI-590K.

\section{More Training and Implementation Details}
\label{sec:supp_training_details}

\subsection{Backbone and Adaptation}
We build \modelname{} on \textbf{Qwen3-VL-8B-Instruct} and apply LoRA to the language-model linear layers.
The visual encoder, aligner, and geometry encoder are frozen throughout both SFT and UC-SSRL.
Unless otherwise stated, all training uses bfloat16 precision with DeepSpeed ZeRO-3.
This design keeps training lightweight and makes UC-SSRL a post-SFT refinement stage rather than full-model retraining.
LoRA uses rank 16 and scaling factor $\alpha{=}32$.
The trainable-parameter ratio is small because only LoRA parameters are updated.

\subsection{SFT Settings}
We instruction-tune the model on a mixture of indoor VSI-590K and outdoor nuScenes-10K using next-token prediction.
The default configuration is:
\begin{itemize}
\item \textbf{Optimization:} AdamW optimizer, learning rate $1\times10^{-4}$, per-device batch size 1, gradient accumulation 2, and 200 training steps.
\item \textbf{LoRA:} rank 16 and scaling factor $\alpha{=}32$.
\item \textbf{Sequence length:} maximum sequence length 5,120 with right truncation.
\item \textbf{Efficiency settings:} FlashAttention, padding-free batching, gradient checkpointing, and multi-process data loading.
\end{itemize}

\subsection{GCM Configuration}
We enable the Geometry-Consistent Memory during both SFT and UC-SSRL.
The memory uses $M{=}64$ memory tokens with dimension $d_{\text{mem}}{=}512$ and $n_{\text{chunks}}{=}4$ chunks.
Memory tokens are injected every 4 transformer layers starting from layer 8.
Retrieval uses fused hierarchical selection with top-$K$ chunk retrieval and top-$K$ frame retrieval, both set to $K{=}8$.
Each chunk contains 8 value tokens, and each selected frame contributes 4 value tokens.
We use FAISS acceleration and cache retrieval keys when possible to reduce repeated retrieval overhead.

\subsection{UC-SSRL Settings}
UC-SSRL is initialized from the SFT checkpoint and updates only LoRA parameters.
We keep the same batch size, step count, and optimizer setting as SFT for controlled comparison.
The overall UC-SSRL weight is set to $1\times10^{-2}$.
The two-view sampler constructs paired contexts by perturbing temporal windows and retrieval contexts while keeping the underlying scene and question unchanged.
This setting optimizes cross-view consistency without requiring additional human annotations.

\subsection{Inference Protocol}
We follow the official evaluation protocol of each benchmark.
For VSI-Bench and OSI-Bench, we use their official MCA and NA evaluation metrics.
For MMSI-Video-Bench, we report both \textit{Sufficient-Coverage} and \textit{Uniform-50} settings.
Unless otherwise stated, all methods use the same decoding parameters and frame-budget settings for fair comparison.
Efficiency and scalability experiments are measured on the same A100 80GB GPU setting.

\section{Visualization}
\label{sec:visualization}

\begin{figure}[t]
    \centering
    \includegraphics[width=\linewidth]{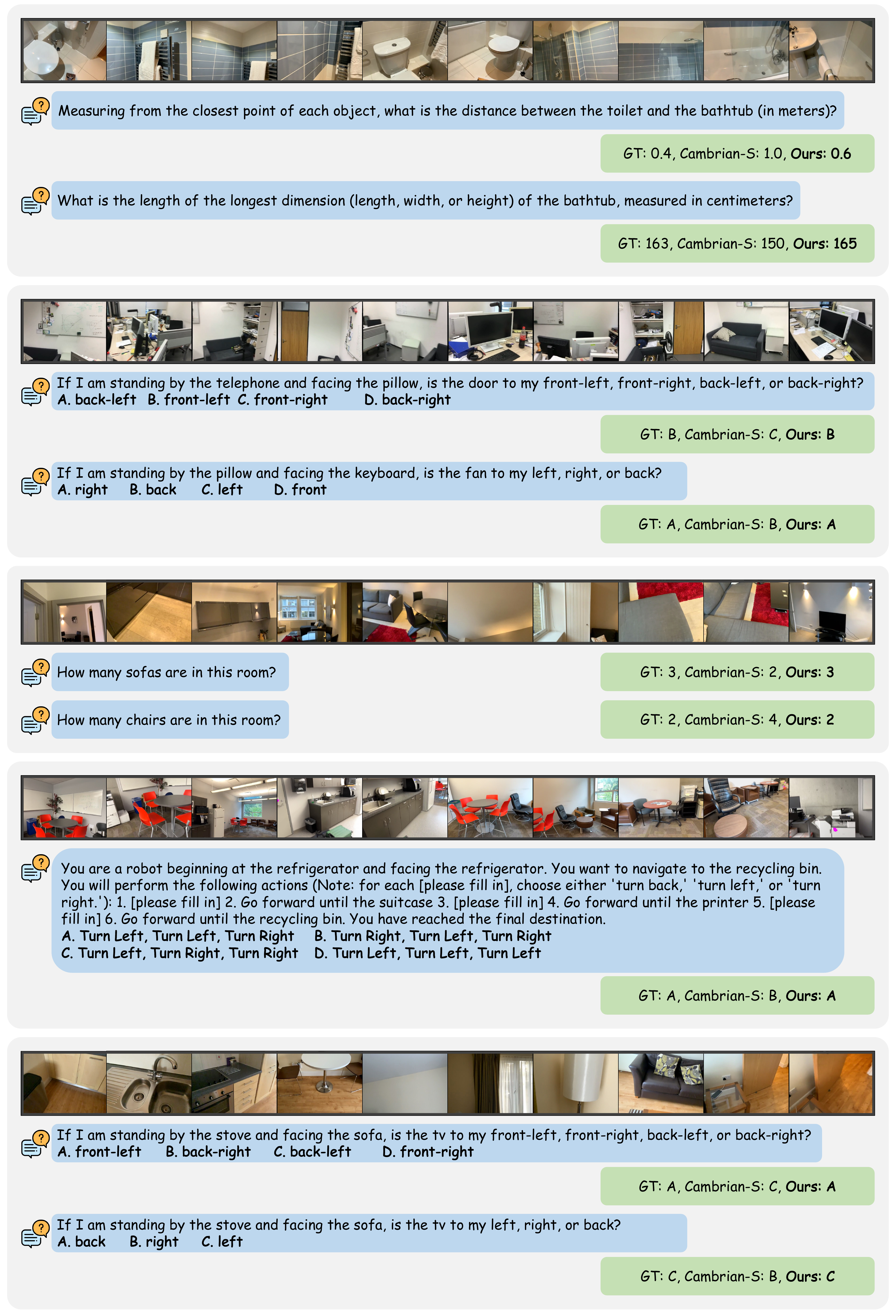}
    \caption{\textbf{Qualitative results on VSI-Bench.}}
    \label{fig:vsi_qual}
\end{figure}

\begin{figure}[t]
    \centering
    \includegraphics[width=\linewidth]{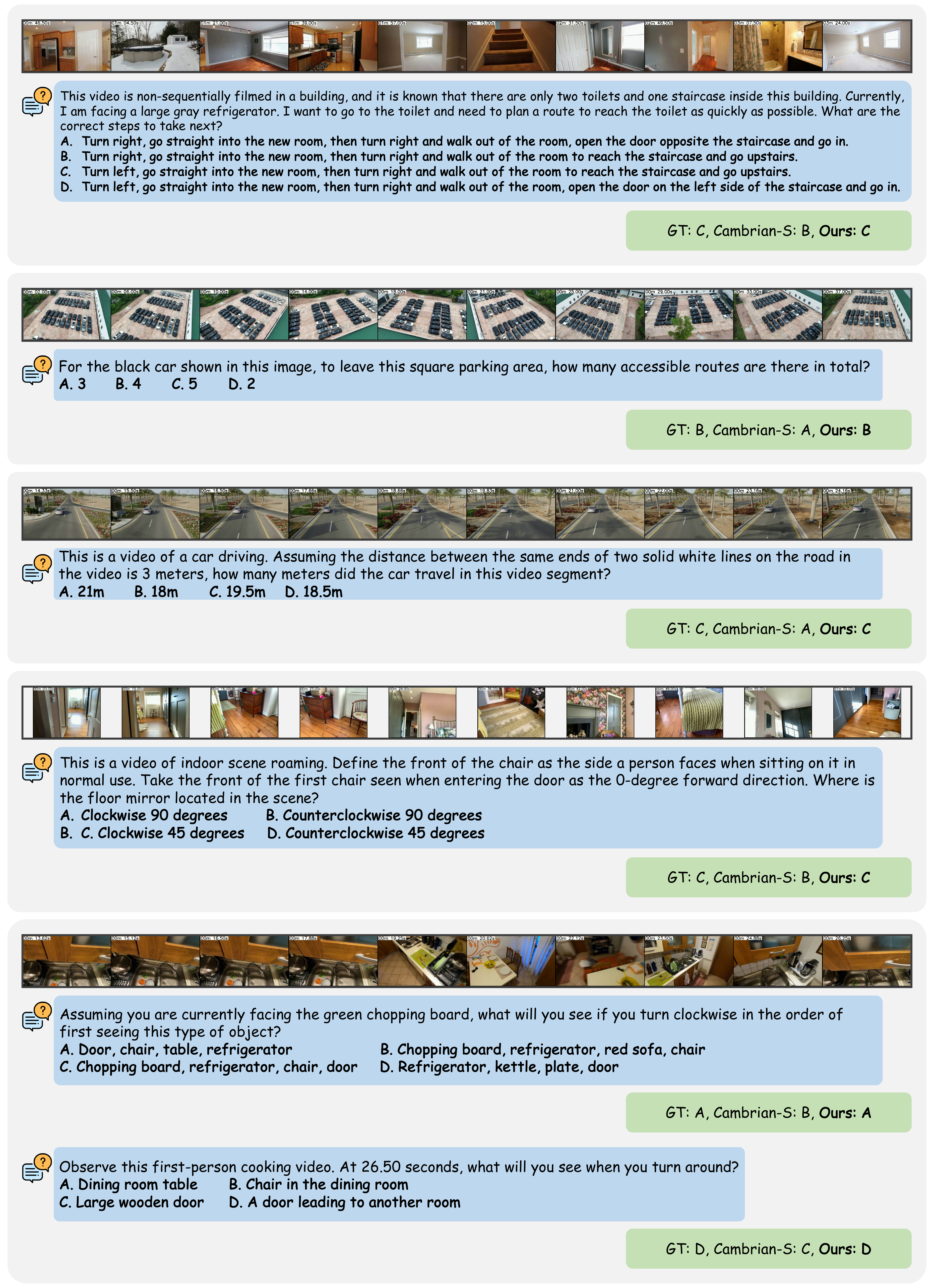}
    \caption{\textbf{Qualitative results on MMSI-Video-Bench.}}
    \label{fig:mmsi_qual}
\end{figure}

\begin{figure}[t]
    \centering
    \includegraphics[width=\linewidth]{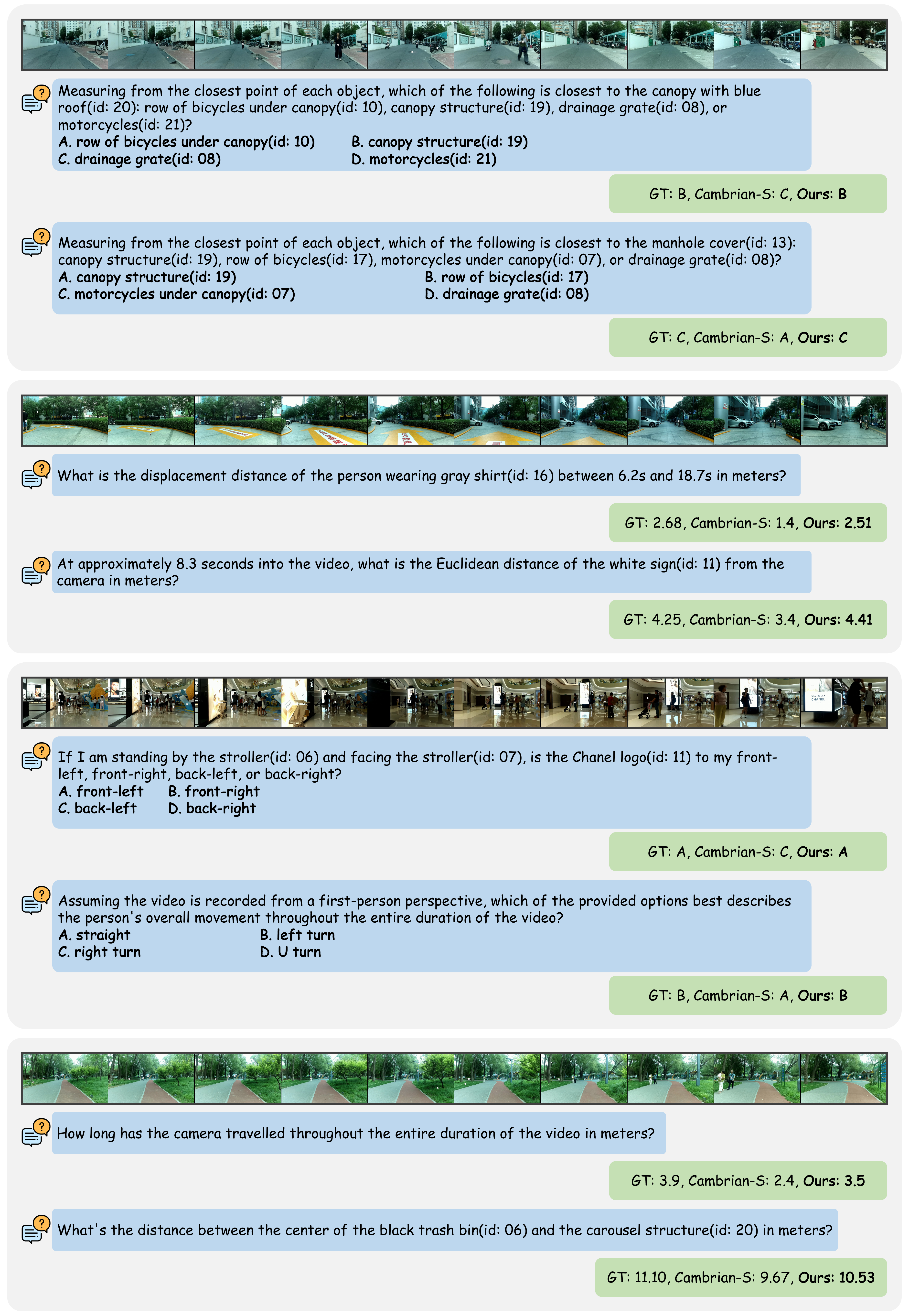}
    \caption{\textbf{Qualitative results on OSI-Bench.}}
    \label{fig:osi_qual}
\end{figure}

Figs.~\ref{fig:vsi_qual}, \ref{fig:mmsi_qual}, and\ref{fig:osi_qual} provide additional qualitative examples on VSI-Bench, MMSI-Video-Bench, and OSI-Bench.
The examples cover metric estimation, relative-direction reasoning, object counting, navigation, and motion-related questions.
Compared with the baseline, \modelname{} produces answers closer to the ground truth across different spatial reasoning settings, supporting the effectiveness of geometry-consistent evidence organization.

\section{Limitation}

While \modelname{} improves efficiency and robustness through geometric consistency, several limitations remain:
\begin{itemize}
\item \textbf{Geometry estimation errors.} \modelname{} uses predicted pose and depth cues for memory control. Errors under motion blur, reflective surfaces, or textureless regions may affect writing, fusion, and retrieval quality.
\item \textbf{Threshold-based memory control.} Although our sensitivity analysis shows stable performance across a range of thresholds, adaptive gating and fusion may further improve robustness across diverse scenes.
\item \textbf{Long-horizon scaling.} Extremely long videos may still require stronger memory compression or learned eviction strategies to maintain bounded compute under strict budgets.
\end{itemize}
Future work may improve geometry estimation, develop adaptive memory control, and extend consistency learning to broader video spatial reasoning settings.

\end{document}